\documentclass[letterpaper, 10 pt, conference]{ieeeconf}  

\IEEEoverridecommandlockouts                              

\usepackage{lipsum}

\usepackage{amsfonts,amssymb,amsmath,bm}

\usepackage{graphicx}
\usepackage{siunitx}
\usepackage{glossaries}

\usepackage{color}
\usepackage{ifthen}
\usepackage{flushend}
\newboolean{authnotes}
\usepackage{tabularx}
\usepackage{multirow}
\usepackage{multicol}
\usepackage{makecell}
\usepackage{booktabs}
\usepackage{comment}
\usepackage[hidelinks]{hyperref}
\usepackage[font=footnotesize,labelfont=bf,tableposition=top]{caption}
\usepackage{cite}

\setboolean{authnotes}{true}

\ifthenelse{\boolean{authnotes}}
{
\newcommand{\todo}[1]{{{\bf\color{magenta} TODO: #1}}}
\newcommand{\draft}[1]{{{\color{black} #1}}}

}
{
\newcommand{\todo}[1]{}
\newcommand{\nf}[1]{}
\newcommand{\db}[1]{}
\newcommand{\st}[1]{}
}

\newcommand\scalemath[2]{\scalebox{#1}{\mbox{\ensuremath{\displaystyle #2}}}}

\usepackage{threeparttable}
\usepackage{enumerate}
\usepackage{comment}

\usepackage{graphicx}
\usepackage{wrapfig}

\usepackage[ruled,vlined,linesnumbered]{algorithm2e}

\usepackage[capitalise]{cleveref} 

\usepackage{glossaries}

\usepackage[resetlabels,labeled]{multibib}
\newcites{Supp}{Supplementary References}

\newcommand{\norm}[1]{\left\lVert#1\right\rVert}

\def\delequal{\stackrel{\text{def}}{=}}

\def\se{\mathfrak{se}}

\def\1{\bm{1}}

\def\RR{\mathbb{R}}

\def\vzero{{\bm{0}}}

\def\vtheta{{\bm{\theta}}}

\def\vf{{\bm{f}}}
\def\vg{{\bm{g}}}
\def\vh{{\bm{h}}}

\def\vq{{\bm{q}}}

\def\vs{{\bm{s}}}
\def\vt{{\bm{t}}}

\def\vv{{\bm{v}}}

\def\vx{{\bm{x}}}

\def\vh{{\bm{h}}}
\def\vz{{\bm{z}}}

\def\vepsilon{{\boldsymbol{\epsilon}}}

\def\vpsi{{\boldsymbol{\psi}}}
\def\vphi{{\boldsymbol{\phi}}}

\def\vtheta{{\boldsymbol{\theta}}}
\def\vtau{{\boldsymbol{\tau}}}

\def\mA{{\bm{A}}}

\def\mH{{\bm{H}}}
\def\mI{{\bm{I}}}
\def\mJ{{\bm{J}}}

\def\mM{{\bm{M}}}

\def\mR{{\bm{R}}}

\def\mSigma{{\bm{\Sigma}}}

\DeclareMathAlphabet{\mathsfit}{\encodingdefault}{\sfdefault}{m}{sl}
\SetMathAlphabet{\mathsfit}{bold}{\encodingdefault}{\sfdefault}{bx}{n}

\def\gD{{\mathcal{D}}}
\def\gE{{\mathcal{E}}}

\def\gJ{{\mathcal{J}}}

\def\gL{{\mathcal{L}}}
\def\gM{{\mathcal{M}}}
\def\gN{{\mathcal{N}}}

\def\gX{{\mathcal{X}}}
\def\gY{{\mathcal{Y}}}
\def\gZ{{\mathcal{Z}}}

\newcommand{\E}{\mathbb{E}}

\DeclareMathOperator*{\argmin}{arg\,min}

\makeatletter
\newcommand{\StatexIndent}[1][3]{%
  \setlength\@tempdima{\algorithmicindent}%
  \Statex\hskip\dimexpr#1\@tempdima\relax}
\makeatother
\newacronym{bc}{BC}{Behavioural Cloning}
\newacronym{il}{IL}{Imitation Learning}
\newacronym{irl}{IRL}{Inverse Reinforcement Learning}
\newacronym{ioc}{IOC}{Inverse Optimal Control}
\newacronym{lfd}{LfD}{Learning from Demonstration}
\newacronym{em}{EM}{Expectation Maximization}
\newacronym{promp}{ProMP}{Probabilistic Movement Primitives}
\newacronym{dmp}{DMP}{Dynamic Movement Primitives}
\newacronym{seds}{SEDS}{Stable Estimator of Dynamical Systems}
\newacronym{gmr}{GMR}{Gaussian Mixture Regressor}
\newacronym{gpr}{GPR}{Gaussian Process Regressor}
\newacronym{lwr}{LWR}{Locally Weighted Regressor}
\newacronym{kmp}{KMP}{Kernelized Movement Primitives}
\newacronym{clf}{CLF}{Control Lyapunov Function}
\newacronym{wsaqf}{WSAQF}{Weighted Sum of Asymmetric Quadratic Function)}
\newacronym{nilc}{NILC}{Neurally Imprinted Lyapunov
Candidate}
\newacronym{clfdm}{CLF-DM}{Control Lyapunov Function-based Dynamic Movements}
\newacronym{iflow}{iFlows}{ImitationFlows}
\newacronym{cnmp}{CNMP}{Conditional Neural Movement Primitives}
\newacronym{tpgmm}{TP-GMM}{Task Parameterized GMM}

\newacronym{mp}{MP}{Movement Primitive}
\newacronym{mpflows}{MPFlows}{Movement Primitive Flows}

\newacronym{gcl}{GCL}{Guided Cost Learning}

\newacronym{mle}{MLE}{Maximun Likelihood Estimation}
\newacronym{sde}{SDE}{Stochastic Differential Equation}
\newacronym{ode}{ODE}{Ordinary Differential Equation}

\newacronym{probs}{ProbS}{Probabilistic Segmentation}
\newacronym{crf}{CRF}{Conditional Random Fields}
\newacronym{ppca}{PPCA}{Probabilistic Principal Component Analysis}
\newacronym{gmcc}{GMCC}{Generalized Multiple Correlation Coeficcient}
\newacronym{hri}{HRI}{Human-Robot Interaction}
\newacronym{ip}{IP}{Interaction Primitives}
\newacronym{hmm}{HMM}{Hidden Markov Model}
\newacronym{cac}{CAC}{Canonical Correlation Coefficient}
\newacronym{rv}{$R_v$}{$R_v$ Coefficient}
\newacronym{dcor}{dCor}{Distance Correlation}
\newacronym{dtw}{DTW}{Dynamic Time Warping}
\newacronym{edr}{EDR}{Edit Distance With Real Penalty}
\newacronym{twed}{TWED}{Time Warp Edit Distance}
\newacronym{r2}{$R^2$}{Coefficient of Determination}
\newacronym{sqp}{SQP}{Successive Quadratic Programming}
\newacronym{rkhs}{RKHS}{Reproducing Kernel Hilbert Space}
\newacronym{icnn}{ICNN}{Input-Convex Neural Network}

\newacronym{pca}{PCA}{Principal Component Analysis}

\newacronym{maf}{MAF}{Masked Autoregressive Flow}
\newacronym{iaf}{IAF}{Inverse Autoregressive Flow}
\newacronym{node}{N-ODE}{Neural ODE}
\newacronym{nsflow}{NSF}{Neural Spline Flows}
\newacronym{cnf}{CNF}{Conditional Normalizing Flows}
\newacronym{ffjord}{FFJORD}{Free-form Jacobian of Reversible Dynamics}

\newacronym{gan}{GAN}{Generative Adversarial Networks}
\newacronym{vae}{VAE}{Variational Autoencoders}
\newacronym{ebm}{EBM}{Energy Based Models}
\newacronym{sbm}{SBM}{Score based Models}
\newacronym{nf}{NF}{Normalizing Flows}

\newacronym{inn}{INN}{Invertible Neural Networks}
\newacronym{mcmc}{MCMC}{Markov Chain Monte Carlo}
\newacronym{ld}{LD}{Langevin Dynamics}

\newacronym{cd}{CD}{Contrastive Divergence}
\newacronym{nce}{NCE}{Noise Contrastive Estimation}
\newacronym{dsm}{DSM}{Denoising Score Matching}

\newacronym{ik}{IK}{Inverse Kinematics}

\newacronym{sdf}{SDF}{Signed Distance Field}
\newacronym{se3dif}{SE(3)-DiF}{SE(3)-GraspDiffusionFields} 
\newacronym{se3dif_o}{SE(3)-DiF}{SE(3)-DiffusionFields}
\glsunset{se3dif}
\newacronym{deepsdf}{DeepSDF}{Deep Signed Distace Field}

\newacronym{emd}{EMD}{Earth Mover Distance}

\newacronym{mlp}{MLP}{Multi Layer Perceptron}
\newacronym{rrt}{RRT}{Rapidly-exploring Random Trees}

\newacronym{relu}{ReLU}{Rectified Linear Unit}

\newacronym{vn}{VN}{Vector Neuron}

\newacronym{icp}{ICP}{Iterative Closest Point}

\title{\LARGE \bf 
SE(3)-DiffusionFields: Learning smooth cost functions for \\
joint grasp and motion optimization through diffusion
}

\author{Julen Urain$^{*1}$, Niklas Funk$^{*1}$, Jan Peters$^{1,2,3,4}$, Georgia Chalvatzaki$^{1}$
\thanks{$*$ Authors contributed equally.}%
\thanks{This work received funding by the DFG Emmy Noether Programme (CH 2676/1-1), by the AICO grant by the Nexplore/Hochtief Collaboration with TU Darmstadt, and the EU project ShareWork.}
\thanks{$^{1}$ Technische Universität Darmstadt (Germany), $^2$ German Research Center for AI (DFKI), $^3$ Hessian.AI, $^4$ Centre for Cognitive Science {\tt\footnotesize\{julen.urain, niklas.funk, jan.peters, georgia.chalvatzaki\} @tu-darmstadt.de}}%
}

\let\oldtwocolumn\twocolumn
\renewcommand\twocolumn[1][]{%
    \oldtwocolumn[{#1}{
    \begin{center}
    \vspace{-0.6cm}
		\includegraphics[width=.99\textwidth, height=3.cm]{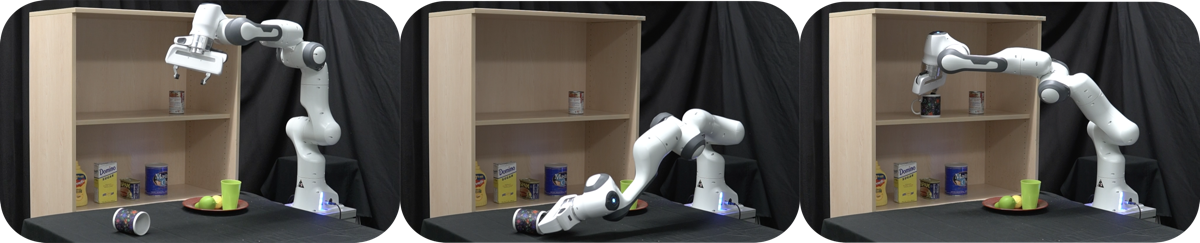}
	\captionof{figure}{Pick and place task in which the robot has to pick a mug and move it to the target pose~(in the shelves) without colliding. We exploit diffusion models for jointly optimizing both grasp and motion and show the successful trajectory from left to right.}
	\label{fig:main_figure}    
        \end{center}
    }]
}
\begin{document}

\maketitle
\thispagestyle{empty}
\pagestyle{empty}

\crefname{appendix}{App.}{apps.}
\crefname{figure}{Fig.}{figs.}
\crefname{section}{Sec.}{secs.}

\begin{abstract}
Multi-objective optimization problems are ubiquitous in robotics, e.g., the optimization of a robot manipulation task requires a joint consideration of grasp pose configurations, collisions and joint limits. While some demands can be easily hand-designed, e.g., the smoothness of a trajectory, several task-specific objectives need to be learned from data. This work introduces a method for learning data-driven SE(3) cost functions as diffusion models. Diffusion models can represent highly-expressive multimodal distributions and exhibit proper gradients over the entire space due to their score-matching training objective. Learning costs as diffusion models allows their seamless integration with other costs into a single differentiable objective function, enabling joint gradient-based motion optimization. In this work, we focus on learning SE(3) diffusion models for 6DoF grasping, giving rise to a novel framework for joint grasp and motion optimization without needing to decouple grasp selection from trajectory generation. We evaluate the representation power of our SE(3) diffusion models w.r.t. classical generative models, and we showcase the superior performance of our proposed optimization framework in a series of simulated and real-world robotic manipulation tasks against representative baselines. Videos, code and additional details are available at:~\url{https://sites.google.com/view/se3dif}
\end{abstract}

\section{INTRODUCTION}
\label{sec:introduction}
Autonomous robot manipulation tasks usually involve complex actions requiring a set of sequential or recurring subtasks to be achieved while satisfying certain constraints, thus, casting robot manipulation into a multi-objective motion optimization problem~\cite{ratliff2009chomp,kalakrishnan2011stomp, schulman2014motion}.
Let us consider the pick-and-place task in \cref{fig:main_figure}, for which the motion optimization should consider the possible set of grasping and placing poses, the trajectories' smoothness, collision avoidance with the environment, and the robot's joint limits. While some objectives are easy to model (e.g., joint limits, smoothness), others (e.g., collision avoidance, grasp pose selection) are more expensive to model and are therefore commonly approximated by learning-based approaches~\cite {rakita2018relaxedik, osa2022motion, mousavian20196, urain2020imitationflows, simeonov2022neural}.

Data-driven models are usually integrated into motion optimization either as sampling functions~(explicit generators)~\cite{mousavian20196, koert2016demonstration}, or cost functions~(scalar fields)~\cite{lambert2022learning, rakita2018relaxedik}. 
When facing multi-objective optimization scenarios, the explicit generators do not allow a direct composition with other objectives, requiring two or even more separate phases during optimization~\cite{murali20206}. Looking back at the example of \cref{fig:main_figure}, a common practice is to learn a grasp generator as an explicit model, sample top-k grasps, and then find the trajectory that, initialized by a grasp candidate, solves the task with a minimum cost. Given the grasp sampling is decoupled from the trajectory planning, it might happen the sampled grasps to be unfeasible for the problem, leading to an unsolvable trajectory optimization problem.
On the other hand, learned scalar fields represent task-specific costs that can be combined with other learned or heuristic cost functions to form a single objective function for a joint optimization process. However, these cost functions are often learned through \textit{cross-entropy optimization}~\cite{mousavian20196, lu2020planning} \textit{or contrastive divergence}~\cite{finn2016guided, lambert2022learning}, creating hard discriminative regions in the learned model that \textit{lead to large plateaus in the learned field with zero or noisy slope regions}~\cite{arjovsky2017towards, miyato2018spectral}, thereby making them unsuitable for pure gradient-based optimization. Thus, it is a common strategy to rely on task-specific samplers that first generate samples close to low-cost regions before optimizing~\cite{mousavian20196, lu2020planning}.

In this work, we propose learning \textit{smooth} data-driven cost functions, drawing inspiration from state-of-the-art diffusion generative models~\cite{song2020score, understanding2022diffusion, huang2022riemannian, de2022riemannian, gnaneshwar2022score}.
By \textit{smoothness}, we refer to the cost function exposing informative gradients in the entire space. 
We propose learning these smooth cost functions in the SE(3) robot's workspace, thus defining task-specific SE(3) cost functions. 
In particular, in this work, we show how to learn diffusion models for 6DoF grasping, leveraging open-source vastly annotated 6DoF grasp pose datasets like Acronym~\cite{eppner2021acronym}.
SE(3)~diffusion models allow moving initially random samples to low-cost regions (regions of good grasping poses on objects) by evolving a gradient-based inverse diffusion process~\cite{song2021train} (cf.~\cref{fig:diffusion}). SE(3) diffusion models come with two benefits. First, we get smooth cost functions in SE(3) that can be directly used in motion optimization. Second, they better cover and represent multimodal distributions, like in a 6DoF grasp generation scenario, leading to better and more sample efficient performance of the subsequent robot planning. 

Consequently, we propose a joint grasp and motion optimization framework using the learned 6DoF grasp diffusion model as cost function and combining it with other differentiable costs (trajectory smoothness, collision avoidance, etc.). All costs combined (learned and hand-designed) form a single, smooth objective function that optimizing it enables the generation of good robot trajectories for complex robot manipulation tasks. This work shows how our framework enables facing grasp generation and classical trajectory optimization as a joint gradient-based optimization loop.

Our contributions are threefold:
(\textbf{1}) we show how to \textbf{learn smooth cost functions in SE(3) as diffusion models}. 
\draft{While score-based  generative modeling has been previously introduced for arbitrary Riemannian manifolds~\cite{de2022riemannian}, we focus on the particular requirements for the Lie group SE(3).}
(\textbf{2}) \textbf{we use the SE(3) diffusion models to learn 6DoF grasp pose distributions as cost functions}. Our experiments show that our learned models generate more diverse and successful grasp poses w.r.t. state-of-the-art grasp generative models.
Once the model is trained, (\textbf{3}) we introduce \textbf{a gradient-based optimization framework for jointly resolving grasp and motion generation}, in which we integrate our learned 6DoF grasp diffusion model with additional task-related cost terms. 
\draft{To properly integrate diffusion models in the motion optimization problem, we rewrite the optimization as an inverse diffusion process, similarly to \cite{janner2022planning}.}
In contrast with previous methods that decouple the grasp pose selection and the motion planning, our framework resolves the grasp and motion planning problem by iteratively improving the trajectory to jointly minimize the learned object-grasp cost term and the task-related costs. We remark that this joint optimization is only possible thanks to the smoothness of our learned diffusion model and using instead a grasp classifier, trained with cross-entropy loss, as cost won't resolve the problem due to its lack of smoothness.
Our quantitative and qualitative results in simulation and the real-world robotic manipulation experiments suggest that our proposed method for learning costs as SE(3) diffusion models enables efficiently finding good grasp and motion solutions against baseline approaches and resolves complex pick-and-place tasks as in \cref{fig:main_figure}.

\section{Preliminaries}
\label{sec:background}
\textbf{Diffusion Models.}
Unlike common deep generative models (\gls{vae}, generative adversarial networks (GAN)) that explicitly generate a sample from a noise signal, diffusion models
learn to generate samples by iteratively moving noisy random samples towards a learned distribution~\cite{song2019generative, song2020score}. A common approach to train diffusion models is by \textsl{\gls{dsm}}~\cite{vincent2011connection, saremi2018deep}.
To apply \gls{dsm} \cite{song2019generative, song2020improved},
we first perturb the data distribution $\scalemath{0.9}{\rho_{\gD}(\vx)}$ with Gaussian noise on $L$ noise scales $\scalemath{0.9}{\gN(\vzero, \sigma_k\mI)}$ with $\scalemath{0.9}{\sigma_1<\sigma_2<\dots<\sigma_L}$,
to obtain a noise perturbed distribution $\scalemath{0.9}{q_{\sigma_k}(\hat{\vx}) = \int_{\vx} \gN(\hat{\vx}|\vx, \sigma_k \mI) \rho_{\gD}(\vx) \textrm{d} \vx}$. To sample from the perturbed distribution, $\scalemath{0.9}{q_{\sigma_k}(\hat{\vx})}$ we first sample from the data distribution $\scalemath{0.9}{\vx\sim\rho_\gD(\vx)}$ and then add white noise $\scalemath{0.9}{\hat{\vx} = \vx + \epsilon}$ with $\scalemath{0.9}{{\epsilon \sim \gN(\vzero, \sigma_k \mI)}}$. Next, we estimate the score function of each noise perturbed distribution $\scalemath{0.9}{\nabla_{\vx} \log q_{\sigma_k}(\vx)}$ by training a noise-conditioned vector field $\scalemath{0.9}{\vs_{\vtheta}(\vx, k)}$, by score matching $\scalemath{0.9}{{\vs_{\vtheta}(\vx, k) \approx \nabla_{\vx}\log q_{\sigma_k}(\vx)}}$ for all $\scalemath{0.9}{k=1,\dots,L}$.
The training objective of \gls{dsm}~\cite{saremi2018deep} is
\begin{align}\label{eq:dsm}
  \scalemath{0.9}{\gL_{\textrm{dsm}} =} &\scalemath{0.9}{\frac{1}{L}\sum_{k=0}^L \E_{\vx, \hat{\vx}}} \scalemath{0.9}{\left[\norm{\vs_{\vtheta}(\hat{\vx}, k) - \nabla_{\hat{\vx}}\log \gN(\hat{\vx}|\vx, \sigma_k^2\mI) }\right],}
\end{align}
with $\scalemath{0.9}{\vx\sim \rho_\gD(\vx)}$ and $\scalemath{0.9}{\hat{\vx} \sim \gN(\vx, \sigma_k \mI)}$
To generate samples from the trained model, we apply Annealed Langevin \gls{mcmc}~\cite{neal2011mcmc}. 
We first draw an initial set of samples from a distribution $\scalemath{0.9}{{\vx_L}\sim \rho_{L}(\vx)}$ and then, simulate an inverse Langevin diffusion process for $\scalemath{0.9}{L}$ steps, from $\scalemath{0.9}{k=L}$ to $\scalemath{0.9}{k=1}$
\begin{align}\label{eq:ld}
  \scalemath{0.9}{
    \vx_{k-1} = \vx_k + \frac{\alpha_k^2}{2} \vs_{\vtheta}(\vx_k, k) + \alpha_k \vepsilon \, ,\, \vepsilon \sim \gN(\vzero, \mI)},
\end{align}
with $\scalemath{0.9}{\alpha_k>0}$ a step dependent coefficient.
Overall, \gls{dsm} \cref{eq:dsm} learns models that output vectors pointing towards the samples of the training dataset $\rho_{\gD}(\vx)$~\cite{song2021train}.
\\
\textbf{SE(3) Lie group.}
The SE(3) Lie group is prevalent in robotics. A point $ \mH = \bigl[ \begin{smallmatrix} \mR & \vt \\
 \vzero & 1
  \end{smallmatrix} \bigr] \in \textrm{SE(3)}$ represents the full pose (position and orientation) of an object or robot link
with $\scalemath{0.9}{\mR \in \textrm{SO(3)}}$ the rotation matrix and $\scalemath{0.9}{\vt \in \RR^3}$ the 3D position. A Lie group encompasses the concepts of group and smooth manifold in a unique body.
Lie groups are smooth manifolds whose elements have to fulfil certain constraints.
Moving along the constrained manifold is achieved by selecting any velocity withing the space tangent to the manifold at $\scalemath{0.9}{\mH}$ (i.e., the so-called tangent space). 
The tangent space at the identity is called \textsl{Lie algebra} and noted $\scalemath{0.9}{\se (3)}$. 
The Lie algebra has a non-trivial structure, but is isomorphic to the vector space $\scalemath{0.9}{\RR^6}$ in which we can apply linear algebra. As in \cite{sola2018micro}, we work in the vector space $\scalemath{0.9}{\RR^6}$ instead of the Lie algebra $\scalemath{0.9}{\se (3)}$.
We can move the elements between the Lie group and the vector space with the logarithmic and exponential maps, $\scalemath{0.9}{\textrm{Logmap}: \textrm{SE(3)} \xrightarrow{} \RR^6$ and $\textrm{Expmap}: \RR^6 \xrightarrow{} \textrm{SE(3)}}$ respectively~\cite{sola2018micro}.
A Gaussian distribution on Lie groups can be defined as
\begin{align}
    \label{eq:lie_gauss_main}
      \scalemath{0.9}{q(\mH | \mH_{\mu}, \mSigma) \propto \exp \left(-0.5~ \norm{\textrm{Logmap}(\mH_{\mu}^{-1} \mH)}_{\mSigma^{-1}}^2  \right)},
\end{align}
with $\scalemath{0.9}{\mH_{\mu}\in SE(3)}$ the mean and $\scalemath{0.9}{\mSigma \in \RR^{6\times 6}}$ the covariance matrix~\cite{chirikjian2014gaussian}. 
This special form is required as the distance between two Lie group elements is not represented in Euclidean space.
Following the notation of \cite{sola2018micro}, given a function $\scalemath{0.9}{f:\textrm{SE(3)} \xrightarrow{} \RR}$, the derivative w.r.t. a SE(3) element, $\scalemath{0.9}{D f(\mH)/ D \mH \in \RR^{6}}$ is a vector of dimension $\scalemath{0.9}{6}$.
We refer the reader to \cite{sola2018micro} and the Appendix in \href{https://sites.google.com/view/se3dif}{project site} for an extended presentation of the SE(3) Lie group. 

\section{SE(3)-DIFFUSION FIELDS}
\label{sec:train}
In this section, we show how to adapt diffusion models to the Lie group SE(3) \cite{sola2018micro}, as it is a crucial space for robot manipulation.
The SE(3) space is not Euclidean, hence, multiple design choices need to be considered for adapting Euclidean diffusion models.
In the following, we first explain the required modifications (\Cref{sec:from_euclidean_to_se3}). Then, we propose a neural network architecture for learning SE(3) diffusion models that represent 6DoF grasp pose distributions and show how we train it~(\cref{sec:architecture}). Finally, we show how to integrate the learned diffusion models into a grasp and motion optimization problem and show how to optimize it jointly considering the grasp and the motion~(\cref{sec:optimization}).
\begin{figure}[t]
	\centering
	\begin{minipage}{.5\textwidth}
	\begin{center}
		\includegraphics[width=.99\textwidth]{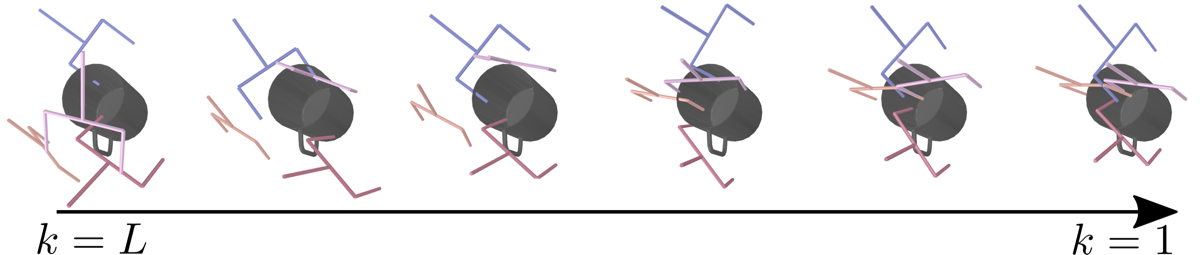}
		\end{center}
	\end{minipage}
	\caption{Generating high quality SE(3) grasp poses by iteratively refining \draft{random} initial samples (k=L) with an inverse Langevin diffusion process over SE(3) elements (\cref{eq:lie_ld}).}
	\label{fig:diffusion}
	\vspace{-.5cm}
\end{figure}
\subsection{From Euclidean diffusion to diffusion in SE(3)}
\label{sec:from_euclidean_to_se3}
A diffusion model in SE(3) is a \textsl{vector field}
that outputs a vector $\vv \in \RR^6$ for an arbitrary query point $\scalemath{0.9}{\mH \in \textrm{SE(3)}}$, i.e., $ \scalemath{0.9}{\vv = \vs_{\vtheta}(\mH, k)}$
with a scalar conditioning variable $k$ determining the current noise scale~\cite{song2019generative}.
\\
\textbf{Denoising Score Matching in SE(3).}
Similar to the Euclidean space version (cf. \cref{sec:background}), \gls{dsm} is applied in two phases. 
We first generate a perturbed data point in SE(3), i.e., sample from the Gaussian on Lie groups~\cref{eq:lie_gauss_main}, $\scalemath{0.9}{\hat{\mH} \sim q(\hat{\mH} |\mH, \sigma_k \mI )}$ with mean $\scalemath{0.9}{\mH\in \rho_{\gD}(\mH)}$ and standard deviation $\scalemath{0.9}{\sigma_k}$ for noise scale $\scalemath{0.9}{k}$.
Practically, we sample from this distribution using a white noise vector $\scalemath{0.9}{\vepsilon \in \RR^6}$,
\begin{align}
    \label{eq:sample_lie_main}
      \scalemath{0.9}{\hat{\mH} = \mH \textrm{Expmap}(\vepsilon) \, ,\, \vepsilon \sim \gN(\vzero, \sigma_k^2\mI)}.
\end{align}
Following the idea of \gls{dsm}, the model is trained to match the score of the perturbed training data distribution.
Thus, \gls{dsm} in SE(3) requires computing the derivatives of the perturbed distribution w.r.t. a Lie group element. Hence, the new \gls{dsm} loss function on Lie groups equates to 
\begin{align}
    \label{eq:lie_dsm}
      \scalemath{0.9}{\gL_{\textrm{dsm}} = \frac{1}{L}\sum_{k=0}^L \E_{\mH, \hat{\mH}}}
      \scalemath{0.9}{\left[\norm{ \vs_{\vtheta}(\hat{\mH}, k) - \frac{D\log q(\hat{\mH}|\mH, \sigma_k \mI)}{D\hat{\mH}} }\right]},
\end{align}
with $\scalemath{0.9}{\mH\sim \rho_\gD(\mH)}$ and $\scalemath{0.9}{\hat{\mH} \sim q(\hat{\mH}|\mH, \sigma_k \mI)}$. Note that, as introduced in \cref{sec:background}, the derivatives w.r.t. a SE(3) element $\scalemath{0.9}{\hat{\mH}}$ outputs a vector on $\scalemath{0.9}{\RR^6}$. In practice, we compute this derivative by automatic differentiation using Theseus~\cite{pineda2022theseus} library along with PyTorch. 
\\
\textbf{Sampling with Langevin \gls{mcmc} in SE(3).}
Evolving the inverse Langevin diffusion process for SE(3) elements (cf. \cref{fig:diffusion} for visualization) requires adapting the previously presented Euclidean Langevin MCMC approach \cref{eq:ld}. In particular, we have to ensure staying on the SE(3) manifold throughout the inverse diffusion process. Thus, we adapt the inverse diffusion in SE(3) as
\begin{align}
    \label{eq:lie_ld}
      \scalemath{0.9}{\mH_{k-1} = \textrm{Expmap}\left(\frac{\alpha_k^2}{2} \vs_{\vtheta}(\mH_k, k) + \alpha_k \vepsilon \right)\mH_k},
\end{align}
with $\vepsilon \in \RR^6$ sampled from $\vepsilon \sim \gN(\vzero, \mI)$ and the step dependent coefficient $\alpha_k>0$. By iteratively applying \cref{eq:lie_ld}, we move a set of \draft{randomly} 
sampled SE(3) poses to the data distribution $\rho_{D}(\mH)$ (See \cref{fig:diffusion}).
\\
\textbf{From the score function to energy model.}
While most of the works in learning diffusion models learn a vector field representing the score $\vs_{\vtheta}$, in our work, we learn a scalar field that represents the energy of the distribution $E_{\vtheta}$. In contrast with learning a score function, learning an \gls{ebm} allow us evaluating the quality of the generated samples and compose it with other cost functions for multi-objective motion optimization. To learn an \gls{ebm} with denoising score matching, we model our score function 
$\vs_{\vtheta}(\mH, k) = - DE_{\vtheta}(\mH, k) / D \mH,$
as the derivative of the \gls{ebm} $E_{\vtheta}$.

\begin{figure*}[t]
	\centering
	\begin{minipage}{.76\textwidth}
		\includegraphics[width=.99\textwidth]{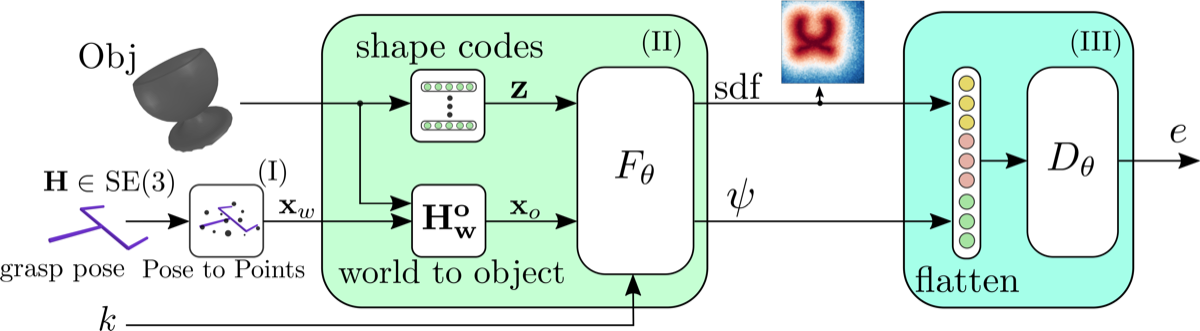}
	\end{minipage}
	\vspace{-0.1cm}
	\caption{\Gls{se3dif}'s architecture for learning 6D grasp pose distributions. 
	We train the model to jointly learn the objects' sdf and to minimize the denoising loss. Given grasp pose $\mH {\in} \textrm{SE(3)}$ we transform it to a set of 3D points $\vx_w {\in} \RR^{N\times3}$ (\textbf{I}). Next, we transform the points into the object's local frame, using the object's pose $\mH_w^o$. Given the resulting points $\vx_o$ and the object's shape code $\vz$ we apply the feature encoder $F_\theta$ (\textbf{II}) to obtain a object and grasp-related features (sdf, $\psi$)$\in\RR^{N\times(\psi+1)}$. Finally, (\textbf{III}) we flatten the features and compute the energy $e$ through the decoder $D_\theta$. \draft{We provide a point-cloud-based implementation in our code repository: \href{https://github.com/TheCamusean/grasp_diffusion}{https://github.com/TheCamusean/grasp\_diffusion}}.
	}
	\label{fig:architecture}
	\vspace{-.7cm}
\end{figure*}
\subsection{Architecture \& training of Grasp SE(3)-DiffusionFields}
\label{sec:architecture}
Even though we can represent any data-driven cost in SE(3) with \gls{se3dif_o}, in this work, we focus on cost functions that capture 6DoF grasp pose distributions conditioned on the object we aim to grasp.
In this work, we assume to have access to the object pose, a reasonable assumption thanks to the impressive results in 6DoF object pose estimation and segmentation \cite{wen2020se}. We defer studying the perception aspect of encoding point clouds into object pose and shape as in~\cite{mousavian20196, jiang2021synergies} for a future work.
We illustrate the architecture for our grasp \gls{se3dif} model in \cref{fig:architecture} and the training pipeline in \cref{alg:training_grasp_dif}.
The proposed model maps an object~(represented by its id and pose) and a 6DoF grasp pose $ \scalemath{0.9}{\mH \in }$~SE(3) to an energy $ \scalemath{0.9}{e \in \RR}$, that measures the grasp quality for the particular object. 

We train the model to jointly match the \gls{sdf} of the object we aim to grasp and predict the grasp energy level by the \gls{dsm} loss~\cref{eq:lie_dsm}. Learning jointly the SDF of the object and the grasp pose improves the quality of the grasp generation \cite{jiang2021synergies, simeonov2021neural}.
During the training, we assume the object's id $\scalemath{0.9}{m}$ and pose $\scalemath{0.9}{\mH_{w}^{o} {\in}}$~SE(3) are available, and we retrieve a learnable object shape code $\vz_m$ given the index $m$ as in \cite{park2019deepsdf}. 
For training the SDF loss, we apply a supervised learning pipeline. Given a dataset of 3D points $\scalemath{0.9}{\vx_w \in \RR^3$ and $\textrm{sdf} \in \RR}$ for a particular object $\scalemath{0.9}{m}$, $\scalemath{0.9}{\gD_{sdf}^m:(\vx_w, \textrm{sdf})}$, we first map the points to the object's reference frame $\scalemath{0.9}{\vx_o=\mH_{w}^{o}\vx_w}$ and then predict the SDF given the feature encoder $\scalemath{0.9}{F_{\vtheta}}$ (See \cref{alg:training_grasp_dif}).

As previously introduced in \cref{eq:lie_dsm}, to apply the \gls{dsm} loss, we compute the energy $\scalemath{0.9}{e \in \RR}$ over the grasp poses $\scalemath{0.9}{\hat{\mH}}$. These grasp poses have been previously obtained by perturbing grasp poses from the dataset $\scalemath{0.9}{\mH\in \rho_{\gD}(\mH)}$ with a noise level $\scalemath{0.9}{k}$~\cref{eq:sample_lie_main}. In our problem, we consider $\scalemath{0.9}{\rho_{\gD}(\mH)}$ to be a distribution of successful grasp poses for a particular object, and learn the energy to approximate the log-probability of this distribution under noise. We compute the energy $\scalemath{0.9}{e}$ given a grasp pose $\scalemath{0.9}{\hat{\mH}}$ in three steps. \textbf{(I)} We transform the grasp pose to a fixed set of $\scalemath{0.9}{N}$ 3D-points around the gripper $\scalemath{0.9}{\vx_g {\in} \RR^{N\times3}}$ in the world frame $\scalemath{0.9}{\vx_w {=} \mH \vx_g}$. We thereby express the grasp pose through a set of 3D points' positions, similar to \cite{simeonov2021neural}. Then, we move the points to the object's local frame, $\scalemath{0.9}{\vx_{o_m} {=} \mH_{w}^{o_m} \vx_w}$. \textbf{(II)} We apply the feature encoding network $\scalemath{0.9}{F_{\vtheta}}$ which is also conditioned on $\scalemath{0.9}{\vz_m}$ and $\scalemath{0.9}{k}$ to inform about the object shape and noise level, respectively. The encoding network outputs both the \gls{sdf} predictions for the query points, $\scalemath{0.9}{\textrm{sdf} {\in} \RR^{N\times1}}$, and a set of additional features $\scalemath{0.9}{\vpsi {\in} \RR^{N\times\psi}}$. 
Thus, the feature encoder's output is of size $\scalemath{0.9}{N {\times} (1+\psi)}$.
\textbf{(III)} We flatten the features and pass them through the decoder $\scalemath{0.9}{D_{\vtheta}}$ to obtain the scalar energy value $\scalemath{0.9}{e}$. Given the energy, we compute the \gls{dsm} loss \cref{eq:lie_dsm}. During training, we jointly learn the objects' latent codes $\scalemath{0.9}{\vz_m}$, and the parameters $\scalemath{0.9}{\vtheta}$ of the feature encoder $\scalemath{0.9}{F_{\vtheta}}$ and decoder $\scalemath{0.9}{D_{\vtheta}}$.
\begin{algorithm}[h]
    \small
	\SetKwInOut{Input}{Given}
	\Input{
		$\vtheta_0$: initial params for $\vz$, $F_{\vtheta}$, $D_{\vtheta}$\;
		Datasets: $\gD_o: \{m, \mH_{w}^{o}\}$, object ids and poses, $\gD_{sdf}^m: \{\vx , \textrm{sdf} \}$, 3D positions $\vx$ and sdf for object $m$, $\gD_{g}^m: \{\mH \}$ succesful grasp poses for object $m$\;}
	\BlankLine
	\For{\scriptsize{$s \leftarrow 0$ \KwTo $S-1$}}{
	    \scriptsize{$k, \sigma_k \leftarrow [0,\dots,L]$}\tcp*{\tiny{sample noise scale}}
		\scriptsize{$m, \mH_{w}^{o} \in \gD_o$}\tcp*{\tiny{sample objects ids and poses}}
		\scriptsize{$\vz = \textrm{shape codes}(m)$}\tcp*{\tiny{get shape codes}}
		\scriptsize{\textbf{SDF train}}\\
		$\vx, \textrm{sdf}\in \gD_{sdf}^m$\tcp*{\tiny{get 3D points and sdf for obj. $m$}}
		$\hat{sdf}, \_ = F_{\vtheta}(\mH_{w}^{o}\vx, \vz, k)$\tcp*{\tiny{get predicted sdf}}
		$L_{\textrm{sdf}} = \gL_{\textrm{mse}}(\hat{sdf}, sdf)$\tcp*{\tiny{compute sdf error}}
		\scriptsize{\textbf{Grasp diffusion train}}\\
		$\mH \sim \gD_{g}^m$\tcp*{\tiny{Sample success grasp poses for obj. $m$}}
        $\vepsilon \sim \gN(\vzero, \sigma_k\mI)$\tcp*{\tiny{sample white noise on $k$ scale}}
        $\hat{\mH} = \mH \textrm{Expmap}(\vepsilon)$\tcp*{\tiny{perturb grasp pose \cref{eq:sample_lie_main}}}
        $\vx_{n}^{o} = \hat{\mH} \vx_{n}$\tcp*{\tiny{Transform to N 3d points (see \Cref{fig:architecture})}}
        $\hat{sdf}_{n}, \vpsi_{n} = F_{\vtheta}(\vx_{n}^{o}, \vz_b, k)$\tcp*{\tiny{get features}}
        $\Psi =\textrm{Flatten}(\hat{sdf}_{n}, \vpsi_{n})$\tcp*{\tiny{Flatten the features}}
        $e = D_{\vtheta}(\Psi)$\tcp*{\tiny{compute energy}}
        $L_{\textrm{dsm}} = \gL_{\textrm{dsm}}(e, \hat{\mH}, \mH, \sigma_k)$\tcp*{\tiny{Compute dsm loss \cref{eq:lie_dsm}}}
        \scriptsize{\textbf{Parameter update}}\\
		$L = L_{\textrm{dsm}} + L_{\textrm{sdf}}$\tcp*{\tiny{Sum losses}}
		$\vtheta_{s+1} = \vtheta_{s} - \alpha \nabla_{\vtheta}L$\tcp*{\tiny{Update parameters}
		}}
	\KwRet{$\vtheta^*$}\;
	
	\caption{Grasp \gls{se3dif} Training \label{alg:training_grasp_dif}}
\end{algorithm}
\vspace{-.65cm}

\subsection{Grasp and motion optimization with diffusion models}
\label{sec:optimization}
\label{sec:grasp_motion_opt}
Given a trajectory $\scalemath{0.9}{\vtau:\{\vq_t\}_{t=1}^T}$, consisting of $\scalemath{0.9}{T}$ waypoints, with $\scalemath{0.9}{\vq_t\in \RR^{d_q}}$ the robot's joint positions at time instant $\scalemath{0.9}{t}$; in motion optimization, we aim to find the minimum cost trajectory~$\scalemath{0.9}{\vtau^{*} = \textstyle \argmin_{\vtau} \gJ(\vtau)= \textstyle \argmin_{\vtau} \textstyle \sum_j \omega_j c_j(\vtau)}$,~where the objective function $\scalemath{0.9}{\gJ}$ is a weighted sum of costs $\scalemath{0.9}{c_j}$, with weights $\scalemath{0.9}{\omega_j>0}$. Herein, we integrate the learned \gls{se3dif} for grasp generation as one cost term of the objective function. It is, thus, combined with other heuristic costs, e.g., collision avoidance or trajectory smoothness.
Optimizing over the whole set of costs enables obtaining optimal trajectories jointly taking into account grasping, as well as motion-related objectives.
This differs from classic grasp and motion planning approaches in which the grasp pose sampling and trajectory planning are treated separately~\cite{lagriffoul2014efficiently}, by first sampling the grasp pose, and, then, searching for a trajectory that satisfies the selected grasp. In classic approaches, given the grasp sampling is decoupled from the trajectory planning, it might happen the sampled grasps to be unfeasible for the problem, leading to an unsolvable trajectory planning problem. We hypothesize that jointly optimizing over both the grasp pose and the trajectory allows us to be more sample efficient w.r.t. decoupled approaches.

Given that the learned function is in SE(3) while the optimization is w.r.t. the robot's joint space, we redefine the cost as $\scalemath{0.9}{c(\vq_t, k) = E_{\vtheta}(\phi_{ee}(\vq_t), k)}$, with the forward kinematics  $\scalemath{0.9}{\phi_{ee}:\RR^{d_q} \xrightarrow{} \textrm{SE(3)}}$ mapping from robot configuration to the robot's end-effectors task space.
To obtain minimum cost trajectories, \textit{we frame the motion generation problem as an inverse diffusion process}.
Using a planning-as-inference view~\cite{botvinick2012planning,levine2018reinforcement,urain2021composable,janner2022planning}, we define a desired target distribution as $\scalemath{0.9}{q(\vtau| k) \propto \exp(-\gJ(\vtau, k))}$.
This allows us to set an inverse Langevin diffusion process that evolves a set of random initial particles drawn from a distribution $\scalemath{0.9}{\vtau_L \sim p_L(\vtau)}$ towards the target distribution $\scalemath{0.9}{q(\vtau| k)}$
\begin{align}
    \label{eq:inverse_diff}
    \scalemath{0.9}{\vtau_{k-1} = \vtau_k + 0.5~\alpha_k^2 \nabla_{\vtau_k} \log q(\vtau|k) +  \alpha_k \vepsilon \,,\, \vepsilon \sim \gN(\vzero, \mI)},
\end{align}
with step dependent coefficient $\scalemath{0.9}{\alpha_k>0}$, noise level moving from $\scalemath{0.9}{k=L}$ to $\scalemath{0.9}{k=1}$, and one particle corresponding to an entire trajectory.
If we evolve the particles by this inverse diffusion process for sufficient steps, the particles at $\scalemath{0.9}{k=1}$, $\scalemath{0.9}{\vtau_1}$ can be considered as particles sampled from $\scalemath{0.9}{q(\vtau| k=1)}$. 
To obtain the optimal trajectory, we evaluate the samples on $\scalemath{0.9}{\gJ(\vtau, 1)}$ and pick the one with the lowest cost.


\section{EXPERIMENTAL EVALUATION}
\label{sec:experiments}

The experimental section is divided in three parts. First, we evaluate our trained model for 6DoF grasp pose generation~(\cref{sec:exp_grasp_gen}).  We train a \gls{se3dif} as a 6DoF grasp pose generative model 
using the Acronym dataset~\cite{eppner2021acronym}. 
This simulation-based dataset contains successful 6DoF grasp poses for a variety of objects from ShapeNet~\cite{chang2015shapenet}.
We focus on the collection of successful grasp poses for 90 different mugs (approximately 90K 6DoF grasp poses). 
We provide a model trained in a larger dataset and conditioned on point cloud in the project page.
We obtain the mugs' meshes from ShapeNet, and train the model as described in \cref{alg:training_grasp_dif}. We generate a set of grasp poses from the learned models and evaluate on successful grasping and diversity.
Second, we evaluate the quality of our trained model when used as an additional cost term for grasp and motion optimization~(\cref{sec:exp_grasp_motion_opt}). We compare the performance of solving a grasp and motion optimization problem jointly (using the learned model as cost function), w.r.t. the state-of-the-art approaches that decouple the grasp selection and motion planning, or heuristically combine them. 
Finally, we validate the performance of our method in a set of real robot experiments (\cref{sec:real_exp}).
\begin{figure}[t]
	\centering
	\begin{minipage}{.5\textwidth}
		\includegraphics[width=.99\textwidth]{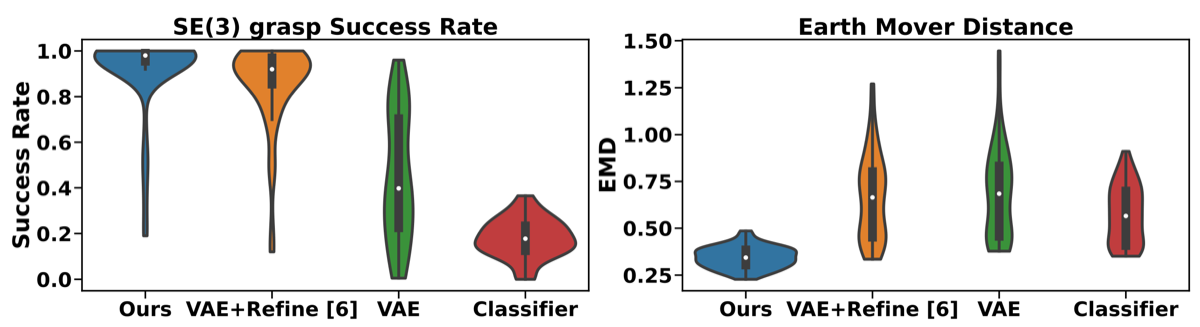}
	\end{minipage}
	\caption{6D grasp pose generation experiment. Left: Success rate evaluation. Right: \gls{emd} evaluation metrics (lower is better).}
	\vspace{-.6cm}
	\label{fig:results_grasp}
\end{figure}
\subsection{Evaluation of 6DoF grasp pose generation}  
\label{sec:exp_grasp_gen}

We evaluate grasp poses generated from our trained grasp \gls{se3dif} model in terms of the success rate, and the \gls{emd} between the generated grasps and the training data distribution. We consider 90 different mugs and evaluate 200 generated grasps per mug. We evaluate the grasp success on Nvidia Isaac Gym~\cite{makoviychuk2021isaac}. The \gls{emd} measures the divergence between two empirical probability distributions~\cite{tanaka2019discriminator}, providing a metric on how similar the generated samples are to the training dataset.
To eliminate any other influence, we only consider the gripper and assume that we can set it to any arbitrary pose.
We generate 6DoF grasp poses from \gls{se3dif} by an inverse diffusion process, following \cref{eq:lie_ld}.
\\
We compare against three baselines. First, based on \cite{mousavian20196, sundermeyer2021contact}, we consider generating grasp poses by first sampling from a decoder of a trained \gls{vae} and subsequently running \gls{mcmc} over a trained
classifier for pose refinement (\gls{vae}+Refine). Second, we consider sampling from the \gls{vae} (without any further refinement). Third, we consider running \gls{mcmc} over the classifier starting from random initial pose~\cite{ten2017grasp}. In this experiment, we assume the object's pose and id/shape to be known, and purely focus on evaluating the models' generative capabilities.
For ensuring a fair comparison, all the baselines consider a shape code $\vz_m$ to encode the object information as presented in \cref{fig:architecture}.
We add a pointcloud-conditioned experiment in the Appendix.
\\
We present the results in \cref{fig:results_grasp}. In terms of success rate, \gls{se3dif} outperforms \gls{vae}+Refine slightly (especially yielding lower variance), and \gls{vae} or classifier on their own significantly. 
The \gls{vae} alone generates noisy grasp poses that are often in collision with the mug. 
In the case of classifier only, the success rate is low. We hypothesize that this might be related with the classifier's gradient, as specifically in regions far from good samples, the field has a large plateau with close to zero slopes~\cite{arjovsky2017towards}. This leads to not being able to improve the initial samples.
Considering grasp diversity, i.e., \gls{emd} metric (lower is better), \gls{se3dif} outperforms all baselines significantly. A reason for the difference, might be that \gls{vae}+Refine overfits to specific overrepresented modes of the data distribution. In contrast, \gls{se3dif}'s samples capture the data distribution more properly.
We, therefore, conclude that \gls{se3dif} is indeed generating high-quality and diverse grasp poses. We add an extended presentation of the experiment in the Appendix in our \href{https://sites.google.com/view/se3dif}{project site}.
\begin{figure}[t]
	\centering
	\begin{minipage}{.5\textwidth}
		\includegraphics[width=.85\textwidth]{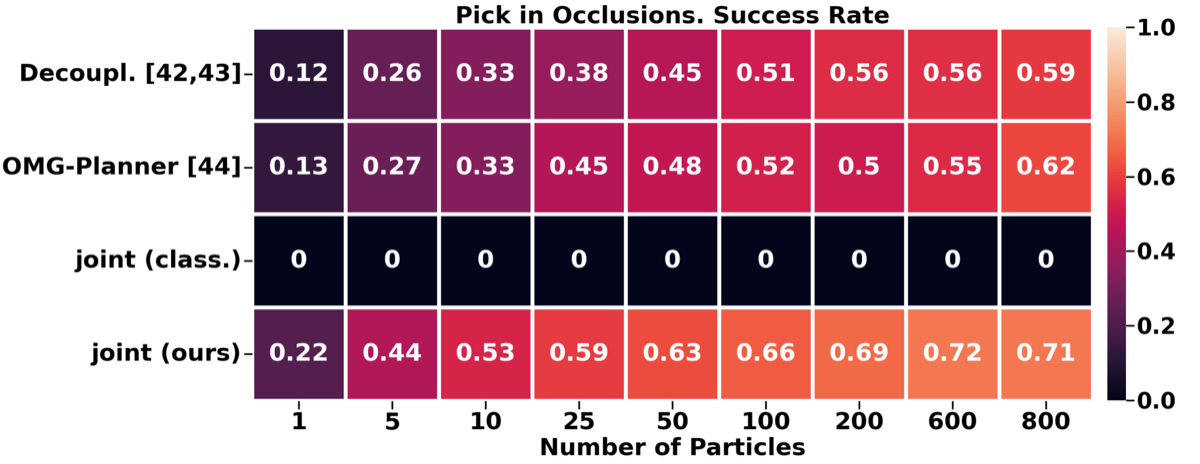}
	\end{minipage}
	\caption{Evaluation Pick in occlusion. We measure the success rate of 4 different methods based on different number of initializations.}
	\vspace{-0.2cm}
	\label{fig:trajopt_results}
	\vspace{-.5cm}
\end{figure}

\subsection{Performance on grasp and motion optimization}
\label{sec:exp_grasp_motion_opt}
\begin{figure}[t]
	\centering
	\begin{minipage}{.5\textwidth}
	    \centering
		\includegraphics[width=.85\textwidth]{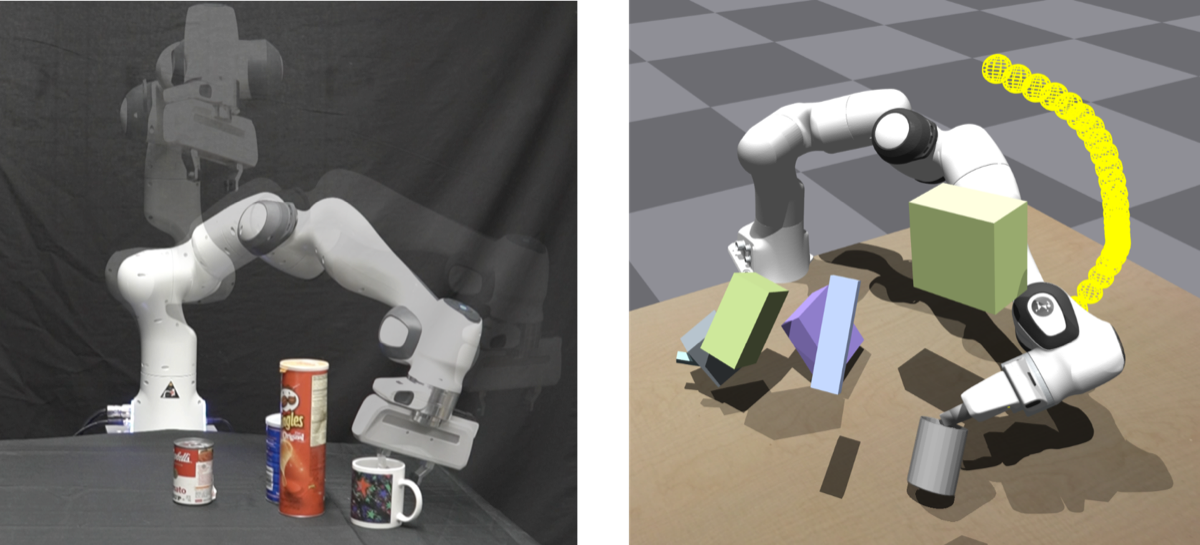}
	\end{minipage}
	\caption{Simulated and real robot environments for picking amidst clutter.}
	\label{fig:robot_envs}
	\vspace{-.6cm}
\end{figure}

We evaluate the performance of our learned grasp \gls{se3dif} as a cost term into multi-objective grasp and motion optimization problems. 
We consider the task of picking amidst clutter (see \cref{fig:robot_envs}) and measure the success rate on solving it. The success is measured based on the robot being able to grasp the object at the end of the execution. In the Appendix in our \href{https://sites.google.com/view/se3dif}{project site}, we provide additional details on the chosen cost functions for the task. 
As introduced in \cref{sec:grasp_motion_opt}, we generate the trajectories by integrating our learned grasp \gls{se3dif} as an additional cost function to the motion optimization objective function. Then, given a set of initial trajectory samples, obtained from a Gaussian distribution with a block diagonal matrix as in \cite{kalakrishnan2011stomp}, we apply gradient descent methods~\cref{eq:inverse_diff} to iteratively improve the trajectories on the objective function.
We evaluate the success rate of the trajectory optimization given a different number of initial samples. As gradient-based trajectory optimization methods are inherently local optimization methods, multiple initializations might lead to better results.
We consider three baselines (see \cref{fig:trajopt_results}). \textbf{Decoupl.}: we adopt the common routing to solve grasp and motion optimization problems in a decoupled way~\cite{rahardja1996vision,  mahler2017dex, murali20206}. We first sample a set of 6DoF grasp poses from a generative model and then plan a trajectory that satisfies the selected grasp pose with CHOMP~\cite{ratliff2009chomp}. Second, we consider the \textbf{OMG-Planner}~\cite{wang2019manipulation}, that applies an online grasp selection and planning approach. Finally, \textbf{joint (class.)}: we consider applying a joint optimization as in our approach, but using a 6DoF grasp classifier as cost function rather than a grasp \gls{se3dif}.
\\
The results in \cref{fig:trajopt_results} present a clear benefit from the joint optimization w.r.t. the decoupled approach and the OMG-Planner. In particular, our proposed joint optimization only requires 25 particles to match the success rate of the decoupled approach with 800 particles. The reason for this significant gap in efficiency is that the decoupled approach generates SE(3) grasp poses that are not feasible given the environment constraints, such as clutter or joint limits. However, when optimizing jointly, we can find trajectories that satisfy all the costs by iteratively improving entire trajectories w.r.t. all objectives. We also observe the importance of using grasp \gls{se3dif} as cost term instead of a grasp classifier. The classifier model lacks proper gradient information to inform how to move the trajectories to grasp the object due to its lack of smoothness in the whole space. Thus, the motion optimization problem is unable to find solutions.





\subsection{Grasp and motion optimization on real robots}
\label{sec:real_exp}
We conducted a thorough real-world evaluation of our joint grasp and motion optimization framework driven by our 6DoF grasp diffusion model, using it as an additional cost function, similarly to our simulated robot manipulation tasks. \cref{fig:main_figure} depicts a sequence of a real-world pick-mug and place-on-shelf scenario. Overall, the experiments aim at assessing the method’s capabilities in realistic conditions
that include, i) non-perfect state information, as the mugs pose is retrieved from an external system (Optitrack) which induces small calibration errors, ii) variations in the mug’s shape, as we use a mug that is slightly different from the one we specify for \gls{se3dif}, and iii) real-world trajectory execution. For optimization, we initialize 800 particles (trajectories), and only execute the one with lowest cost.
\\
In the simplest testing scenario, where the robot has to pick up a mug from various poses in a scene without any clutter, we achieve \textbf{100\%} (20 successes / 20 trials) pickup-success. We also find that our method transfers well to the more difficult scenarios of picking up mugs that are initially placed upside down with \textbf{90\%} (18/20) success, picking in occluded scenes with \textbf{95\%} (19/20) success, and having to pick and place the mug in a desired pose inside the shelf of \cref{fig:main_figure} with \textbf{100\%} (20/20) success. Our real-world results underline the effectiveness of our joint optimization approach. Videos of the experiments also showcase that our method still comes up with very versatile solutions\footnote{Videos in \url{https://sites.google.com/view/se3dif}}. 
Note that we attribute the increased real-world performance w.r.t. the simulated one to the simpler designed experimental scene, i.e., in simulation we considered flying obstacles that were not realizable in the real scene (\cref{fig:robot_envs}).
Nevertheless, our results confirm that our proposed approach is highly performant in real settings, without suffering sim2real discrepancies.

\noindent\textbf{Limitations} In our experiments, we focused on evaluating our diffusion model's performance in grasp generation, besides full trajectory optimization, assuming full object state knowledge, without relying on complex perception systems. Potential sim2real gaps w.r.t. the real environment could potentially arise from imperfect perception, and hand-designed cost terms that may not capture well the relevant task description in more complex scenarios.
Moreover, a limitation comes with increasing number of cost terms, as it becomes more difficult to weight them.



\section{RELATED WORK}
\label{sec:related_work}
\draft{\noindent \textbf{Diffusion models in Robotics} Diffusion models have appeared in Robotics in various tasks, from text-conditioned scene rearrangement \cite{kapelyukh2022dall, liu2022structdiffusion}, decision-making \cite{janner2022planning, ajay2022conditional, wang2022diffusion} and controllable traffic generation\cite{zhong2022guided}. We additionally highlight earlier works like \cite{park2005diffusion}, where a diffusion process is integrated into a motion planning problem.} 
\\
\noindent \textbf{6D grasp generation.} 
6D grasp pose generation is solved with a myriad of methods from classifiers to explicit samplers.
\cite{lou2021collision,ten2017grasp,liang2019pointnetgpd} sample candidate grasps and score them with learned classifiers. 
\cite{yan2018learning} predicts grasping outcomes using a geometry-aware representation.
Contrary to methods classifying grasps, generative models can be trained to generate grasp poses from data \cite{mousavian20196} 
but might require additional sample refinement. 
While the generator in \cite{sundermeyer2021contact} considers possible collisions in the scene, \cite{hager2021graspme} proposes to learn a grasp distribution over the object's manifold.
\cite{jiang2021synergies} uses scene representation learning to learn grasp qualities and explicitly predict 3D rotations. Recently, \cite{weng2022neural} proposed learning a 6 DoF SDF to represent grasp pose generation as a smooth cost function and optimize on top of it.
\\
\noindent \textbf{Integrated grasp and motion planning.}
Due to the interdependence of the selected grasp pose with the robot motion, multiple efforts have tried to integrate both variables into a single planning problem~\cite{dragan2011learning, berenson2011task, vahrenkamp2010integrated, wang2019manipulation, funk2021benchmarking}. In \cite{dragan2011learning, berenson2011task}, goal sets representing grasp poses are integrated as constraints in a motion optimization problem. In \cite{vahrenkamp2010integrated, fontanals2014integrated}, Rapidly-exploring Random Trees~\cite{lavalle1998rapidly} is combined with a TCP attractor to bias the tree towards good grasps. 


\section{CONCLUSION}
\label{sec:conclusion}

We proposed SE(3)-DiffusionFields (SE(3)-DiF) for learning task-space, data-driven cost functions to enable robotic motion generation through joint gradient-based optimization over a set of combined cost functions. 
At the core of SE(3)-DiFs is a diffusion model that provides informative gradients across the entire space and enables data generation through an inverse Langevin dynamics diffusion process.
Besides having demonstrated that SE(3)-DiF generates diverse and high-quality 6DoF grasp poses, we also drew a connection between motion generation and inverse diffusion. Thus, we presented a joint gradient-based grasp and motion optimization framework, which outperforms traditional decoupled optimization approaches. 
Our extensive experimental evaluations reveal the superior performance of the proposed method w.r.t. efficiency, adaptiveness, and success rates.
In the future, we want to explore diffusion models for reactive motion control and the composition of multiple diffusion models to solve complex manipulation tasks in which multiple hard-to-model objectives might arise.

\clearpage
\bibliographystyle{IEEEtran}
\bibliography{bibliography.bib}


\clearpage
\onecolumn
\setlength{\parindent}{10pt}
\begin{appendices}

\section{Theory on SE(3) Lie group: derivatives and distributions}
\label{app:lie_theory}

The Lie group SE(3) is prevalent in robotics. A point $\mH \in \textrm{SE(3)}$ represents the full pose (position and orientation) of an object or robot link
\begin{align}
    \label{se(3)}
    \mH = \begin{bmatrix}
  \mR & \vt \\
 \vzero & 1
  \end{bmatrix} \in \textrm{SE(3)}
\end{align}
with $\mR \in \textrm{SO(3)}$ the rotation matrix and $\vt \in \RR^3$ the 3D position. A Lie group is both a group and a differentiable manifold~(See \cite{sola2018micro} for additional details on groups). Given SE(3) is a differentiable manifold, for any point $\mH \in SE(3)$, there exists a tangent space centered around $\mH$ that is locally diffeomorphic to SE(3). The tangent space can be afterwards map to a Cartesian vector space $\RR^6$. In particular, the tangent space at identity is known as Lie algebra and is noted by $\se (3)$.  

We can interact between the Lie group and the Lie algebra through the logmap and expmap functions.
The logmap is a function that maps a point $\mH\in \textrm{SE(3)}$ to the Lie algebra $\se (3)$, $\textrm{logmap}:\textrm{SE(3)}\xrightarrow{} \se (3)$. Inversely, 
the expmap moves the points $\vh^{\wedge} \in \se(3)$  to the Lie group $\textrm{SE(3)}$, $\textrm{expmap}: \se (3) \xrightarrow{} \textrm{SE(3)}$. 
Additionally, we can relate the elements in the Lie algebra $\se (3)$ with the Cartesian vector space $\RR^6$ through the hat and vee functions. The hat function $(\cdot)^{\wedge}: \RR^6 \xrightarrow{} \se (3)$ maps the points in the vector space $\vh \in \RR^6$ to the Lie algebra $\se (3)$. Inversely, the vee function $(.)^{\vee}: \se(3) \xrightarrow{} \RR^6$, moves the points in the Lie algebra $\vh^{\wedge} \in \se(3)$ to the vector space $\RR^6$. The vector space $\RR^6$ is isomorphic to $\se (3)$. Then, we can move any point from $\se (3)$ to $\RR^6$ and back. Nevertheless, $\vh \in \RR^6$ representation is more useful in our case as we can apply Linear algebra on them. Finally, we additionally call Logmap the map from SE(3) to $\RR^6$, $\textrm{Logmap} = \textrm{logmap}(.)^{\vee}: \textrm{SE(3)} \xrightarrow{} \RR^6$ and Expmap the map from $\RR^6$ to SE(3), $\textrm{Expmap} = \textrm{expmap}(.^{\wedge}):  \RR^6 \xrightarrow{} \textrm{SE(3)}$. Note that we use the upper case~(Logmap, Expmap) to represent a mapping to the vector space and the lower case~(logmap, expmap) to represent the mapping to the Lie algebra.

A vector field ${f: \textrm{SE(3)} \xrightarrow{} \RR^6}$ is a function that outputs a vector in the Cartesian vector space $\RR^6$ for any point in SE(3). The vector's values are dependent on a particular tangent space centered at $\mH \in \textrm{SE(3)}$.
Given that there exist infinite tangent spaces~(one per point in SE(3)), the value of the vectors might vary depending on the tangent space. We can transform the vectors related with one tangent space to another, with the adjoint matrix operator. The adjoint matrix operator is a linear map $\dot{\vh}_{1} = \mA_{0}^{1} \dot{\vh}_{0}$, that transform a vector $\dot{\vh}_0 \in \RR^6$ tied with  the tangent space centered at $\mH_0 \in \textrm{SE(3)}$ to the vector $\dot{\vh}_1 \in \RR^6$ tied with the tangent space centered at $\mH_1 \in \textrm{SE(3)}$.

\subsection{Derivatives on Lie groups}
\label{app:lie_derivative}
To properly define derivatives on Lie groups, we are required to consider the geometry of the manifold. Given a function $\vf(\cdot): \RR^m\xrightarrow{} \RR^n$, the Jacobian is defined as
\begin{align}
    \label{eq:jac_euclid}
    \mJ = \frac{\partial f(\vx)}{\partial \vx} \delequal \lim_{\vtau\xrightarrow{}\vzero} \frac{f(\vx + \vtau) - f(\vx)}{\vtau} \in \RR^{n\times m},
\end{align}
with $\vtau \in \RR^m$.
Nevertheless, if we aim to compute the Jacobian on the SE(3) Lie group, we are required to adapt the formulation, as we cannot directly sum $\vx$ and $\vtau$. Given a function $f(\cdot):\gM \xrightarrow{} \gN$ from the manifold $\gM$ to the manifold. $\gN$, the Jacobian is defined as
\begin{align}
    \label{eq:deriv_lie}
    \frac{Df(\gX)}{D \gX} \delequal \lim_{\vtau \xrightarrow{}\vzero} \frac{f( \vtau \oplus \gX) \ominus f(\gX)}{\vtau} 
    = \lim_{\vtau \xrightarrow{}\vzero} \frac{\textrm{Logmap}(f(\gX)^{-1} f(\textrm{Expmap}(\vtau)\gX))}{\vtau} \in \RR^{m\times n},
\end{align}
where $m$ is the dimension of the manifold $\gM$ and $n$, the dimension of the manifold $\gN$. $\gX \in \gM$ is an element in $\gM$ and the output $f(\gX) \in \gN $ an element in $\gN$.
The plus $\oplus$ and minus $\ominus$ operators must be selected appropriately: $\oplus$ for the domain $\gM$ and $\ominus$ for the codomain $\gN$~\cite{sola2018micro}.
In our work, we derive assuming the left Jacobian~\eqref{eq:deriv_lie}; yet as presented in \cite{sola2018micro}, it is also possible to compute the right Jacobian.
For the case of SE(3), the Jacobian of the function $f$ will transform a vector of dimension $n$ to a vector in $\RR^6$.
Similarly, to functions mapping between Euclidean spaces, we can apply the chain rule given functions that map between manifolds. Given $\gY = \vf(\gX)$ and $\gZ = \vg(\gY)$, the Jacobian of $\gZ = \vg(\vf(\gX))$ is defined 
\begin{align}
    \mJ = \frac{D\gZ}{D \gX} = \frac{D \gZ}{D \gY} \frac{D \gY}{D \gX} =  \frac{D (\vg(\gY))}{D \gY} \frac{D (\vf(\gX))}{D \gX},
\end{align}
by the concatenation of the Jacobians of each function.

\subsection{Distributions on Lie groups}
To apply the score matching loss, we first sample a datapoint from a Gaussian distribution $q_{\sigma_k}(\hat{\vx}|\vx) = \gN(\hat{\vx}|\vx, \sigma_k \mI)$ with the mean $\vx \sim \rho_{\gD}(\vx)$ sampled from the data distribution. 
A sample from $q_{\sigma_k}(\hat{\vx}|\vx)$ can be easily obtain by perturbing a datapoint from the demonstrations with white noise $\hat{\vx} = \vx + \epsilon$ with $\epsilon \sim \gN(\vzero, \sigma_k\mI)$.
Nevertheless, given SE(3) is not an Euclidean space, we cannot directly sample from a Gaussian distribution as the generated sample might fall out of the manifold. In our work, we adapt the Gaussian distribution to Lie Groups. Similarly to \cite{chirikjian2014gaussian}, we model the sampling distribution in SE(3) as
\begin{align}
    \label{eq:lie_dist}
    q_{\sigma_k}(\hat{\mH}|\mH) \propto \exp \left(-\frac{1}{2} \norm{\textrm{Logmap}(\mH^{-1}\hat{\mH})}^2_{\Sigma^{-1}}\right),
\end{align}
where $\Sigma = \sigma_k^2 \mI$ the covariance matrix. Following the intuition from \cite{chirikjian2014gaussian}, as long as $\sigma_k$ is small enough, the tails of the distribution decay to zero along every geodesic path leading away from identity. 
We can sample from \eqref{eq:lie_dist}, 
\begin{align}
\hat{\mH} = \mH\textrm{Expmap}(\vepsilon)\, ,\, \vepsilon\sim \gN(\vzero,\sigma_k \mI),
\end{align}
by first converting a white noise sample to a SE(3) element, and then, perturbing the mean of the distribution $\mH$ with transformed white noise. 

\subsubsection{Score function in SE(3)}
The score matching loss encourages the gradient of the parameterized model $D E_{\vtheta}(\mH)/D\mH$ to match the score of the perturbed distribution \eqref{eq:lie_dist}. We call $\vphi = \textrm{Logmap}(\mH^{-1}\hat{\mH})$ and $\mM = \mH^{-1}\hat{\mH}$. We compute the score function by the chain rule
\begin{align}
    \frac{D \log  q_{\sigma_k}(\hat{\mH}|\mH)}{D \hat{\mH}} = \frac{D (-\frac{1}{2}\norm{\vphi}_{\Sigma^{-1}}^{2})}{D\vphi} \frac{D(\textrm{Logmap}(\mM))}{D \mM} \frac{D(\mH^{-1}\hat{\mH})}{D \hat{\mH}}.
\end{align}
The first part can be directly computed in the Euclidean space
\begin{align}
    \frac{D (-\frac{1}{2}\norm{\vphi}_{\Sigma^{-1}}^{2})}{D\vphi} =  \frac{\partial (-\frac{1}{2}\norm{\vphi}_{\Sigma^{-1}}^{2})}{\partial \vphi} = -\frac{\vphi}{\sigma_k^2}
\end{align}
that is the score function of a Gaussian distribution on Euclidean spaces. This is the score that is matched in \cite{song2019generative}. The second term
\begin{align}
    \frac{D(\textrm{Logmap}(\mM))}{D\mM} = \mJ_l^{-1}(\vphi)
\end{align}
is the inverse left-Jacobian on SE(3)~(See \cite{sola2018micro}). The third term
\begin{align}
    \frac{D(\mH^{-1}\hat{\mH})}{D \hat{\mH}} = \textbf{Adj}_{\mH^{-1}}
\end{align}
is the adjoint over $\mH^{-1}$.

\section{Algorithmic details}
\label{app:alg_details}

In this section, we provide the pseudocode for all 3 main algorithms (for training the diffusion models, for generating grasp poses, and for handling motion optimization with diffusion).

\subsection{Algorithmic implementation of the training procedure}
\label{app:training_pipeline_pseudocode}

\begin{algorithm}[h]
    \small
	\SetKwInOut{Input}{Given}
	\Input{
		$\vtheta_0$: initial parameters of the function $E_{\vtheta}$\;
		$S$: Optimization steps\;
		Dataset $\gD:\{\{\mH_{k,i}\}_{i=0}^{I_k}, \{\vx_{k,j} , \textrm{sdf}_{k,j} \}_{j=0}^{J_k} , o_k   \}_{k=1}^{K}$: $K$ objects, $J$ 3D points $\vx_{k,j} \in \RR^3$  per object with the $\textrm{sdf}_{k,j}\in \RR$ SDF value for each point. $I$ $\mH_{k,i} \in SE(3)$ good grasp poses per object, $o_k\in \RR$ object id.\;
		$\mH_{w}^o = \mI$: object's pose set to identity for training.
		}
	\BlankLine
	\For{$s \leftarrow 0$ \KwTo $S-1$}{
		$o_b \in \gD$\tcp*{Sample a minibatch of $b$ objects ids}
		$\vz_b = \textrm{shape codes}(o_b, \vtheta_{s})$\tcp*{get all shape codes (see \Cref{fig:architecture})}
		$\vx_{b,j}, \textrm{sdf}_{b,j}\in \gD$\tcp*{SDF: Sample a minibatch of $j$ 3D points and sdf per object}
		$\hat{sdf}_{b,j}, \_ = F_{\vtheta}(\vx_{b,j}, \vz_b, k)$\tcp*{SDF:get predicted sdf (see \Cref{fig:architecture})}
		$l_{\textrm{sdf}} = \gL_{\textrm{mse}}(\hat{sdf}_{b,j}, sdf_{b,j})$\tcp*{SDF:compute sdf error}
		$\mH_{b,i} \sim \gD$\tcp*{DIF: Sample a minibatch of $i$ grasp poses per object}
		$k, \sigma_k \leftarrow [0,\dots,L]$\tcp*{DIF:Sample a noise level}
        $\vepsilon_{b,i} \sim \gN(\vzero, \sigma_k\mI)$\tcp*{DIF:sample a white noise}
        $\hat{\mH}_{b,i} = \mH_{b,i} \textrm{Expmap}(\vepsilon_{b,i})$\tcp*{DIF:perturb grasp poses \cref{eq:sample_lie_main}}
        $\vx_{b,i,n}^{o} =\mH_{w}^o \hat{\mH}_{b,i} \vx_{n}$\tcp*{DIF:Transform N 3d points (see \Cref{fig:architecture})}
        $\hat{sdf}_{b,i,n}, \vphi_{b,i,n} = F_{\vtheta}(\vx_{b,i,n}^{o}, \vz_b, k)$\tcp*{DIF:get latent features (see \Cref{fig:architecture})}
        $\Phi_{b,i}=\textrm{Flatten}(\hat{sdf}_{b,i,n}, \vphi_{b,i,n})$\tcp*{DIF:Flatten the features (see \Cref{fig:architecture})}
        $e_{b,i} = D_{\vtheta}(\Phi_{b,i})$\tcp*{DIF:Compute energy (see \Cref{fig:architecture})}
        $l_{\textrm{dsm}} = \gL_{\textrm{dsm}}(e_{b,i}, \hat{\mH}_{b,i}, \mH_{b,i}, \sigma_k)$\tcp*{DIF:Compute dsm loss with \cref{eq:lie_dsm}}
		$l = l_{\textrm{dsm}} + l_{\textrm{sdf}}$\tcp*{Sum losses}
		$\vtheta_{s+1} = \vtheta_{s} - \alpha \nabla_{\vtheta}l$\tcp*{Update parameter $\vtheta$}
		}
	\KwRet{$\vtheta^*$}\;
	
	\caption{Training procedure for \gls{se3dif} \label{alg:training}}
\end{algorithm}

\Cref{alg:training} summarizes the training procedure for obtaining our \gls{se3dif} diffusion models. This section thus complements \Cref{sec:architecture}. Before starting to explain the training procedure, we want to point out that we are dealing with a combined objective (line 16). 
On the one hand, we refine the representation to learn the object's sdf.
Learning the sdf should instill geometric reasoning into the proposed architecture.
In the algorithm, all operations that are related with this part are commented with "SDF".
On the other hand, our model should also be capable to match the score of the perturbed data distribution. Therefore, we use the denoising score matching loss as the second objective. All operations related to diffusion are marked with "DIF".

In every training iteration, we first sample a minibatch of $b$ object ids. Next, we query the shape codes for all the selected objects. Please note that the shape codes are also learnable parameters, and actually updated during the training procedure. 
Afterwards follow the typical steps for learning the sdfs' of the selected object, i.e., sampling j 3D points per object and their groundtruth sdf values, before querying the predictions by the network and constructing the loss function.
From line 7 on follow the steps for score matching. 
We first start by sampling i grasp poses per object from the Acronym dataset \cite{eppner2021acronym}. The dataset originally contains good (successful) and bad (unsuccessful) grasping poses. However, as we are only interested in learning the distribution of successful grasping poses, we do not consider the bad ones in this sampling step.
Next follow the steps of sampling noise and perturbing the previously selected grasping poses (lines 8-10).
Following our explanations in \Cref{sec:architecture}, we represent the SE(3) grasping poses through a collection of N 3D points that are sampled around the gripper's pose.
We subsequently query our architecture to receive the features for each of these 3D points representing the grasping pose.
Subsequently, we combine all of these points corresponding to a single grasping pose through flattening to obtain the predicted grasp quality (i.e., energy).
We finally compute the \gls{dsm} loss function, add the two objectives and perform gradient descent to update our network's parameters as well as the object's shape codes.

\subsection{Algorithmic implementation of 6D grasp generation using SE(3)-DiF}
\label{app:grasp_generation_pseudocode}

In \Cref{alg:optimization_se3} we provide pseudocode for SE(3) grasp generation, closely following \Cref{sec:from_euclidean_to_se3}.
We nevertheless want to point out that in some experiments in which also table collisions have to be considered, $E_\vtheta$ might consist of multiple terms and therefore not only represent the energies output by our learned diffusion model.

\begin{algorithm}[h]
    \small
	\SetKwInOut{Input}{Given}
	\Input{$\{\sigma_k\}_{k=1}^{L}$: Noise levels\;
		$L$: Diffusion steps\;
		$\epsilon$: step rate\;
		Initialize $n_s$ initial samples $\mH_{L}^{n_s} \sim p_{L}(\mH)$ 
		}
	\BlankLine
	\For{$k \leftarrow L$ \KwTo $1$}{
	    $e_{n_s} = E_\vtheta(\mH_{k}^{n_s}, k)$\tcp*{Compute the energy per $\mH_{k}^{n_s}$}
	    $\alpha_k = \epsilon \cdot \sigma_{k}/\sigma_L$\tcp*{Select step size $\alpha_k$}
	    $\vepsilon \sim \gN(\vzero, \mI)$\tcp*{Sample white noise vector of size $\mathbb{R}^6$}
	    $\mH_{k-1}^{n_s} = \textrm{Expmap}\left( -\frac{\alpha_k^2}{2} \frac{De_{n_s}}{D\mH_{k}^{n_s}} + \alpha_k \vepsilon \right)\mH_{k}^{n_s} ,$\tcp*{Make a ld step}
		}
	\KwRet{$\mH_0^{n_s}$}\;
	\caption{SE(3) grasp generation pipeline \label{alg:optimization_se3}}
\end{algorithm}

\subsection{Algorithmic implementation of robot trajectory optimization using SE(3)-DiF}
\label{app:robot_trajectory_pseudocode}

\Cref{alg:optimization_traj} summarizes the procedure for trajectory optimization using inverse diffusion. 
The pseudocode follows \Cref{sec:grasp_motion_opt}.
Again, the total cost per trajectory is usually a combination of multiple cost terms.
Finally, we only return the minimum cost trajectory and execute it in simulation / on the real robot.

\begin{algorithm}[h]
    \small
	\SetKwInOut{Input}{Given}
	\Input{$\{\sigma_k\}_{k=1}^{L}$: Noise levels\;
		$L$: Diffusion steps\;
		$T$: trajectory lenght\;
		$Q$: dimension of the configuration space\;
		$\epsilon$: step rate\;
		Initialize $n_s$ initial samples $\vtau_{L}^{n_s} \sim p_{L}(\vtau)$ of size $Q \times T$ each
		}
	\BlankLine
	\For{$k \leftarrow L$ \KwTo $1$}{
	    $c_{n_s} = \gJ(\vtau_{k}^{n_s}, k)$\tcp*{Compute the total cost per $\vtau_k^{n_s}$}
	    $\alpha_k = \epsilon \cdot \sigma_{k}/\sigma_L$\tcp*{Select step size $\alpha_k$}
	    $\vepsilon \sim \gN(\vzero, \mI)$\tcp*{Sample white noise vector of size $\mathbb{R}^{Q\times T}$}
	    $\vtau_{k-1}^{n_s} = \vtau_k^{n_s} + \frac{1}{2}\alpha_k^2 \nabla_{\vtau_k} c_{n_s} +  \alpha_k \mathbf{\vepsilon}$\tcp*{Make a ld step}
		}
	\KwRet{$\argmin_{\vtau_{0}^{n_s}} \gJ(\vtau_{0}^{n_s}, 0)$}\;
	
	\caption{Trajectory optimization pipeline \label{alg:optimization_traj}}
\end{algorithm}

\section{Extended experimental evaluation}
\label{app:ext_experiments}

\subsection{Evaluation of SE(3)-DiffusionFields as 6D grasp pose generative models}
\label{app:exp_grasp_eval}

\begin{figure*}[t]
	\centering
	\begin{minipage}{.9\textwidth}
		\includegraphics[width=.99\textwidth]{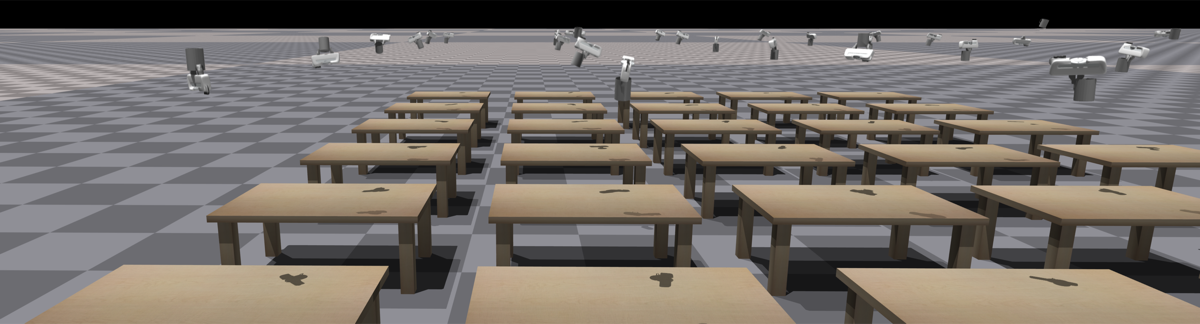}
	\end{minipage}
	\caption{A frame of the grasp success evaluation for the model \gls{se3dif} in Nvidia Isaac Sim.}
	\label{fig:grasp_success_nvidia}
\end{figure*}

In the following, we provide an extended presentation of the experiment in \Cref{sec:exp_grasp_gen}. We measure the success rate with the physics simulator Nvidia Isaac Sim. We present a visualization of the evaluation environment in \Cref{fig:grasp_success_nvidia}. To evaluate the success rate of each model, we first generate $200$ SE(3) grasp poses with each model and for each object. As we can observe in \Cref{fig:grasp_success_nvidia}, the generated grasps are diverse and consider multiple grasping points. For our model, we get the initial SE(3) elements by sampling from a normal distribution on Lie groups
\begin{align}
    \mH_0 = \textrm{Expmap}(\vepsilon) \,,\, \vepsilon \sim \gN(\vzero, \sigma \mI)
\end{align}
with $\sigma = \sigma_K$, the biggest noise level during the training.
Then, we evaluate the grasps quality in Nvidia Isaac Gym. We reset the Franka's end effector in the chosen grasp pose. We smoothly close the fingers until a tight grip is achieved and lift the gripper to a certain height. We consider the grasp to be successful if, after the lift, the mug remains close to the gripper. We also evaluate the divergence of the generated samples distribution w.r.t. data distribution. This divergence informs about how well the learned distribution matches the data distribution, covering all modes. We measure this divergence with the \gls{emd}~\cite{tanaka2019discriminator}. We first sample $N=1000$ grasp poses from the data distribution and from the learned model, respectively, and build a table with the relative distance between all the SE(3) grasp poses as
\begin{align}
    d_{\textrm{SO(3)}  + \RR^3}(\mH_i, \mH_j) = \norm{\vt_i - \vt_j} +  \norm{\textrm{LogMap}(\mR_i^{-1} \mR_j)},
\end{align}
with $\vt_i$ and $\vt_j$ the 3D position and $\mR_i$ and $\mR_j$ the rotation matrix of $\mH_i$ and $\mH_j$ respectively. Then, we solve a Linear Sum Assignment optimization problem~\cite{crouse2016implementing}. This problem solves an optimal transport problem that will search for the least-distance one-to-one assignment between the samples in the data distribution and the sampled grasp poses from the learned model. The smaller the distance, the closer the generated samples are from the data distribution.

We compare the performance of \gls{se3dif} w.r.t. three models that are inspired by 6dof-GraspNet~\cite{mousavian20196} and present the results in \Cref{fig:results_grasp}. We have trained a \gls{vae} to generate 6D grasp poses and a classifier to discriminate between good and bad grasp poses. The classifier network shares the same architecture of \gls{se3dif}, proposed in \Cref{fig:architecture}. For the \gls{vae}, we have trained a conditioned \gls{vae} that receives as input the shape code of the object to grasp and the 6D pose. We jointly train a \gls{deepsdf}~\cite{park2019deepsdf} that shares the shape code with the conditioned \gls{vae}.
We trained the classifier with a cross-entropy loss and added a gradient regularizer to encourage smoother gradients. Nevertheless, when the grasp poses are too far from the data distribution, the classifier lacks informative gradients that would allow us to move the grasp poses to the high-probability regions~(See \Cref{fig:results_grasp}).

\subsection{Evaluation of SE(3)-DiffusionFields for robot grasp pose generation}
\label{app:rechability_eval}

\begin{figure*}[t]
	\centering
	\begin{minipage}{.99\textwidth}
		\includegraphics[width=.99\textwidth]{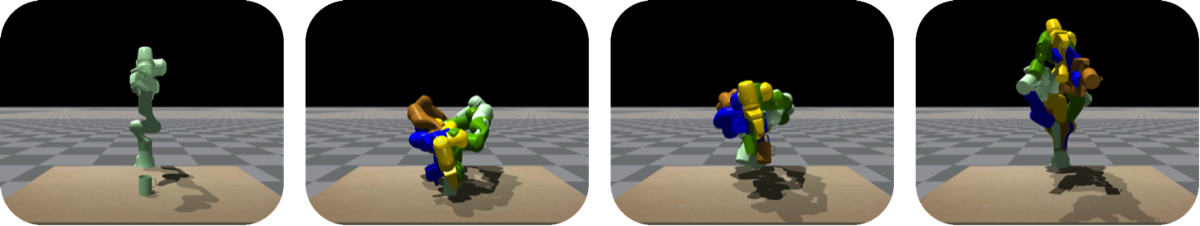}
	\end{minipage}
	\caption{Visualization of evaluation procedure for robot grasp pose generation.  Note that the pictures illustrate the evaluation of the 5 lowest cost particles using our proposed joint optimization with \gls{se3dif}. Importantly, each particle is evaluated in its own environment (environment is identified by colored arm \& mug) and there are no collisions between different environments. Left to right: 1) All environments start with the same initial mug \& robot pose. 2) Setting the arms to the optimized robot grasp pose. 3 \& 4) Attempting to lift the mugs. In this case, all the particles result in successes.}
	\label{fig:robot_grasp_pose_gen}
\end{figure*}

This section complements the findings presented in \Cref{sec:exp_grasp_motion_opt}. 
The results for the robot grasp pose generation have also been obtained in Nvidia Isaac Gym, using the procedure as shown in \Cref{fig:robot_grasp_pose_gen}. 

For obtaining the results, the optimizations not only consider the grasp pose in SE(3), but also the robot joint configuration. In particular, for the two end-end approaches, i.e., classifier \& joint opt, we aim to minimize the following objective function 
\begin{align}
    \gJ(\vq) = E_{\vtheta}(\phi_{ee}(\vq)) + c_{\textrm{table coll.}}(\vq)
\end{align}
with the learned grasp costs $E_{\vtheta}$ and the table collision cost. 
The \textbf{table collision avoidance cost} is computed for all the collision spheres in the robot ${\vx_{c}=(x_{c}, y_{c}, z_{c}) \in \RR^3}$. Given the radius for a particular collision body is $r_c \in \RR$
\begin{align}
    c_{\textrm{table coll.}}(\vq) = \sum_{c=0}^{K} \textrm{ReLU}(-(z_{c} -z_{\textrm{table}} - r_{c})) .
\end{align}

For the separate optimization procedure, we first only optimize for the grasp poses $\mH_{\textrm{grasp pose}}$ through
\begin{align}
    \gJ(\mH_{\textrm{grasp pose}}) = E_{\vtheta}(\mH_{\textrm{grasp pose}}),
\end{align}
thereby not taking into account the current pose of the object, nor any other environmental constraint and thus have to subsequently optimize the following cost function in joint space (for fixed $\mH_{\textrm{grasp pose}}$)
\begin{align}
    \gJ(\vq) = c_{\textrm{des grasp dist}}(\phi_{ee}(\vq), \mH_{\textrm{grasp pose}}) + c_{\textrm{table coll.}}(\vq)
\end{align}
with current grasping pose $\mH(\vq)$ and the cost on the distance to the previously optimized grasp pose
\begin{align}
    c_{\textrm{des grasp dist}}(\mH(\vq), \mH_{\textrm{grasp pose}}) = 10 \norm{\vt_q - \vt_{\textrm{grasp pose}}} +  \norm{\textrm{LogMap}(\mR_q^{-1} \mR_{\textrm{grasp pose}})}.
\end{align}
Note the additional factor of 10 due to the different scales of position error in $[m]$ and orientation error in $[rad]$.

As we have shown in the main paper (cf. \Cref{sec:exp_grasp_motion_opt}), and as it is also underlined by the accompanied videos~(accesible in the linked website in the abstract), using the classifier, and the separate optimization performs substantially worse performance compared to our proposed joint optimization using \gls{se3dif}.

For the separate optimization procedure (sample + opt), we actually even ran two variants. 
The results of optimizing ten joint configuration samples per previously sampled grasp pose ($n_{sm}{=}10$) have been shown in the main paper, thus the second optimization phase even considers 1000 samples in total (100 grasping poses $\times$ 10 samples per grasp pose). When comparing these results to only optimizing one joint configuration per grasp in the second stage ($n_{sm}=1$), we observe that optimizing for finding the single desired grasp pose while also avoiding table collisions is difficult.
Allowing 10 joint
\begin{table}[t]
\vspace{-10pt}
\begin{center}
\caption{Comparing different approaches for robot grasp pose generation with $n_s=100$ initial samples.}
\scalebox{0.75}{
\label{table:robo_grasp_pose_generation1_app}
\begin{tabular}{l|cc|cc}
       & \multicolumn{2}{c|}{Objects upright} & \multicolumn{2}{c}{Objects flipped} \\
 Method          & $s_\Omega$ & $s_1$ & $s_\Omega$ & $s_1$ \\
\hline
\hline

joint opt (classifier) & 0.03 & 0.77 & 0.03 & 0.68 \\
sample (\gls{se3dif}) + opt ($n_{sm}=1$)    &0.11 & 0.79 &  0.03 &  0.57 \\
sample (\gls{se3dif}) + opt ($n_{sm}=10$)    &0.46 & 0.80 &  0.12 &  0.76 \\
\textbf{\textbf{Ours}} & \textbf{0.62} & \textbf{0.88} & \textbf{0.49} & \textbf{0.88} \\
\hline

\end{tabular}
}
\end{center}
\vspace{-10pt}
\end{table}
configuration samples per grasp pose and only evaluating the best one ($n_{sm}{=}10$) performs substantially better, but still worse compared to our proposed joint optimization with \gls{se3dif}. Particularly, the experiments showcase a performance drop w.r.t. the flipped mug scenario. This underlines the major shortcoming of not being adaptive w.r.t. the current environment. The split optimization for grasp pose and joint configuration results in many proposed grasp poses which are simply infeasible. Contrarily, for our proposed joint optimization the ratio of overall successful particles $s_\Omega$ drops only slightly, and $s_1$ even remains on the same high level of $0.88$.
We, thus, conclude that end-end gradient-based optimization with our \gls{se3dif} model results in highly performant, reliable, and adaptive robot grasp pose generation, despite the multi-objective scenario.

\textbf{Exact weighting of cost terms}

In \Cref{tab:robo_grasp_pose_generation_costs}, we additionally present the exact weighting of the cost terms that have been used for generating the results presented in \Cref{sec:exp_grasp_motion_opt} \& \Cref{app:rechability_eval}.

\begin{table*}[h]
\begin{center}
\scalebox{0.7}{
\begin{tabular}{l|c|c|c|c|c}
      & & \multicolumn{4}{c}{Weighting of cost terms} \\
      & & \multicolumn{3}{c}{} & Minimize distance \\
      & & \multicolumn{2}{c|}{Grasp cost} & Table collision & to desired grasp \\
 Method & Phase  & $E_{\vtheta}(\mH_{\textrm{grasp pose}})$ & $E_{\vtheta}(\phi_{ee}(\vq))$   & $c_{\textrm{table coll.}}(\vq)$ & $c_{\textrm{des grasp dist}}(\phi_{ee}(\vq), \mH_{\textrm{grasp pose}})$ \\
\hline
\hline

joint opt (classifier) &  & - & 2 & 3 & - \\
\hline
\textbf{our} joint opt (\gls{se3dif}) &  & - & 2 & 3 & - \\
\hline
\hline
\hline
sample (\gls{se3dif}) & 1) grasp sampling & 1 & - &  - & - \\
+ opt  & 2) opt robot pose & - & - & 0.5 & 1.5 \\
\hline

\end{tabular}
}
\end{center}
\caption{This table summarizes the weighting of the individual cost terms that have been used for generating robot grasp poses, as presented in \Cref{sec:exp_grasp_motion_opt} \& \Cref{app:rechability_eval}. The table's first two rows describe the weighting of the cost terms when running one single joint optimization procedure in which grasp generation and avoiding table collisions are considered jointly. While for the method in the first row, we use the trained classifier as described in \Cref{sec:exp_grasp_motion_opt}, all the other approaches use our proposed \gls{se3dif} diffusion model as the cost function for evaluating grasp poses. Moreover, table's last two rows detail the weighting of the cost terms for the split, two-stage optimization procedure. Note that the separate optimization thus requires running two optimizations.}
\label{tab:robo_grasp_pose_generation_costs}
\end{table*}

\subsection{Evaluation of SE(3)-DiffusionFields for joint grasp and motion optimization}
\label{app:trajectory_eval}
In the following, we provide an extended presentation of the experiments in \Cref{sec:exp_grasp_motion_opt}. We evaluate the performance of \gls{se3dif} as cost function in a trajectory optimization problem. We consider three robot tasks in which both the selection of the grasping pose, and the trajectory planning are required. We explore if we can use \gls{se3dif} as a cost in a single trajectory optimization problem and jointly optimize both for the trajectory and the grasp pose at the last waypoint. The objective function
\begin{align}
    \gJ(\vtau, k) = E_{\vtheta}(\phi_{ee}(\vq_{t=T}), k) + \sum_k c_k(\vtau)
\end{align}
is composed of both the learned \gls{se3dif}, $E_{\vtheta}$ and a set of heuristics cost functions that represent different subtasks (trajectory smoothness, collision avoidance, \dots).
All trajectories are planned in the configuration space.
Then, we frame the optimization problem as an inverse diffusion process and diffuse a set of initial trajectory samples as presented in \Cref{sec:grasp_motion_opt}. We sample the initial trajectories as straight trajectories in the configuration space towards a randomly sampled configuration.
After diffusing a set of trajectories, we pick the one with the lowest accumulated cost $\gJ(\vtau, 1)$. We evaluate this approach in three tasks that require the planning of both the trajectory and the grasping pose: picking an object with occlusions, picking and reorienting an object and pick and placing on shelves.

\subsubsection{Picking with occlusions}
\label{app:pick_occlu}
\begin{figure*}[h]
	\centering
	\begin{minipage}{.99\textwidth}
		\includegraphics[width=.99\textwidth]{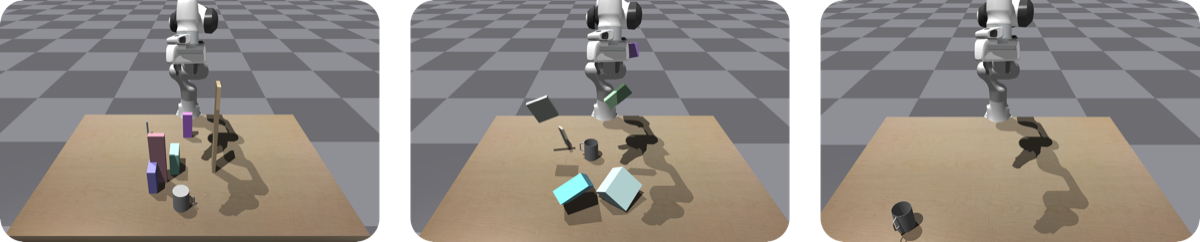}
	\end{minipage}
	\caption{Scenarios for Picking with occlusions. The boxes and the table are obstacles and the robot must find a trajectory to grasp the mug. We consider the mug might be positioned both upright and upside-down.}
	\label{fig:pick_occlusion_scenarios}
\end{figure*}

In the following, we provide an extended presentation of the experiment on  picking an object with occlusions. The experimental evaluation is performed in three different scenarios, with the mug initialized both in normal pose or upside down. We illustrate the scenarios in \Cref{fig:pick_occlusion_scenarios}.
We evaluate the success for 100 trajectories with different mugs positions for the three environments. The success is measured by following the generated trajectory. Once the robot is in the last position of the generated trajectory, we close the fingers. We consider a success case if the mug is in contact with both fingers once the fingers close.

The objective function for this problem is defined by the following cost functions: (a) Grasp Pose \gls{se3dif} over the final configuration $\vq_T$, (b) a trajectory smoothess cost, (c) a table avoidance cost, (d) box avoidance costs, (e) initial configuration fixing cost and, (f) a pregrasp cost. Given some of the costs are defined in the task space, we use a differentiable robot kinematic model, based on Facebook's kinematics model~\cite{pmlr-v120-sutanto20a}. We define the collision body of our robot by a set of spheres similar to \cite{pmlr-v164-bhardwaj22a}. We set the \textbf{trajectory smoothness cost}
\begin{align}
    c_{\textrm{smooth}}(\vtau) = \sum_{t=1}^{T-1} \norm{\vq_{t+1} - \vq_t}^2
\end{align}
as the minimization of the relative distance between the neighbour points in the trajectory. This cost can be thought as a spring making all the point in the trajectory be attracted between each other. The \textbf{table collision avoidance cost} is computed for all the collision spheres in the robot ${\vx_{ct}=(x_{ct}, y_{ct}, z_{ct}) \in \RR^3}$. Given the radius for a particular collision body is $r_c \in \RR$
\begin{align}
    c_{\textrm{table coll.}}(\vtau) = \sum_{t=1}^{T} \sum_{c=0}^{K} \textrm{ReLU}(-(z_{ct} -z_{\textrm{table}} - r_{c}))
\end{align}
with $z_{\textrm{table}}$ the height of the table and a \gls{relu} to bound the cost. Given we have access to the \gls{sdf} of the collision obstacles in the environment, we can set the \textbf{box collision cost} as
\begin{align}
    c_{\textrm{box coll.}}(\vtau) = \sum_{t=1}^{T} \sum_{c=0}^{K} \textrm{ReLU}(- (\textrm{SDF}(\vx_{ct})- r_{c})).
\end{align}
In the trajectory optimization problems, we might want to fix the initial configuration to the current robot configuration. While the easiest approach is not updating $\vq_0$ during optimization, we can alternatively set a \textbf{initial configuration fixing cost}
\begin{align}
    c_{fix}(\vtau) = \norm{\vq_1 - \vq_{\textrm{init}}}
\end{align}
with $\vq_{\textrm{init}}$ the initial configuration of the robot. Finally, we set also a \textbf{pregrasping cost}. It is common to approximate to the grasp in the cartesian space from a grasp a few centimeters over the grasp pose. We set a cost that encourages the optimized trajectory to approximate in this way
\begin{align}
    c_{\textrm{pregrasp}}(\vtau) = \sum_{t=T-n}^{T-1} d_{\textrm{SO(3)}  + \RR^3}(\mH_{ee,t}, \mH_{\textrm{pre,t}})
\end{align}
with $\mH_{ee,t}$ the end effector pose in the instant $t$ and $\mH_{\textrm{pre,t}} = \mH_{ee, T}\mH_{z,t}$ a pose that is to a certain distance over the $z$ axis from the final pose. 

\begin{figure*}[t]
	\centering
	\begin{minipage}{.99\textwidth}
		\includegraphics[width=.99\textwidth]{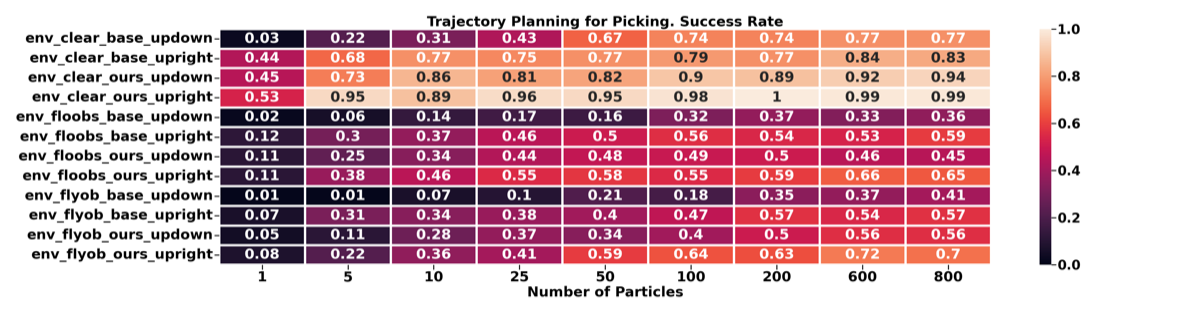}
	\end{minipage}
	\caption{Evaluation of the Success for picking with occlusions.}
	\label{fig:pick_occlusion_evaluation_total}
\end{figure*}

As baseline, we also evaluate the performance of solving the task in a hierarchical approach. In this case, we first sample a SE(3) grasp pose given our learned \gls{se3dif} and then, we solve the trajectory optimization problem, given the target grasp pose is fix with the cost $d_{\textrm{SO(3)}  + \RR^3}$. The complete evaluation can be found in \Cref{fig:pick_occlusion_evaluation_total}. 

We evaluate the success rate of the model assuming a different set of initial particles. We observe that the performance in all the cases increases when considering more initial particles. Gradient based motion optimization is an inherently locally optimization method and therefore, its performance is highly influenced by the initialization. To enhance the performance, multiple initial particles, initialized in different states might explore better the optimization field and find more optimal solutions. We observe that the joint optimization approach outperforms the hierarchical approach in all the cases. This is expected solution. A hierarchical approach decouples the grasp selection from the trajectory optimization. Then, if the selected grasp is unfeasible for the robot, we will not be able to find a good trajectory. Instead a joint optimization problem iteratively updates the trajectory improving both the grasp cost and the rest of the costs. Therefore, we find that jointly optimizing is more sample efficient than a hierarchical approach. 

\textbf{Exact weighting of cost terms}

In \Cref{tab:pick_occlu_weights} we present the weighting of the individual cost terms that we have used to obtain the trajectories for these scenarios of having to pick the mug under occlusions.

\begin{table}[h]
\begin{center}
\scalebox{0.7}{
\begin{tabular}{c|c|c}
Description & Cost & Weight \\
\hline
\hline
Grasp pose evaluation & $E_{\vtheta}(\phi_{ee}(\vq_{T}), k)$ & $.5$\\
\hline
Trajectory smoothness & $c_{\textrm{smooth}}(\vtau)$ & $10.$ \\
\hline
Table collision avoidance & $c_{\textrm{table coll.}}(\vtau)$ & $20.$ \\
\hline
Box collision cost (other obstacles) & $c_{\textrm{box coll.}}(\vtau)$ & $20.$\\
\hline
Initial configuration fixing cost &  $c_{fix}(\vtau)$ & $10.$ \\
\hline
Pregrasping cost & $c_{\textrm{pregrasp}}(\vtau)$ & $5.$ \\

\hline

\end{tabular}
}
\end{center}
\caption{This table summarizes the weighting of the individual cost terms that have been used for generating robot trajectories for the task of picking a mug under occlusions, as presented in \Cref{sec:exp_grasp_motion_opt} \& \Cref{app:pick_occlu}.}
\label{tab:pick_occlu_weights}
\end{table}

\subsubsection{Pick and reorient}
\label{app:pick_reorient}
In the following we provide an extended presentation of the experiment of picking and reorienting an object. This experiment aims to explore the performance of \gls{se3dif} in a complex manipulation task as the one of picking an object and reorient it. We highlight that the whole optimization problem on how to grasp the object, and how to move it to a target pose is solved in a single optimization loop. The problem is interesting as the optimized trajectory should not only consider that there is a collision free path to an affordable grasp pose, but also, that the chosen grasp pose allows us to put the object in a desired target pose. We evaluate the performance of our model 100 times in which the objects are initialized in an arbitrary random pose and have to be placed in an arbitrary placing pose. We consider a trial to be successful, if after executing the whole trajectory, the distance between the grasped object and target pose is smaller to a threshold.

The objective function for this problem maintains multiple costs from the pick on occluded problem (trajectory smoothness, pregrasp, initial target fix, table collision). Additionally, we consider the grasp \gls{se3dif} at the instant $t=T/2$ (we aim to grasp an object in the middle of the trajectory). Finally, we want to impose that the relative position of the object with respect to the gripper in the grasping moment should be the same as the relative position in the placing. We impose this by first computing the pose in the object's frame $\mH_{ee,t}^{o} = (\mH_{o,t}^{w})^{-1} \mH_{ee,t}^{w} $, with $\mH_{ee,t}^{w}$ being the end effector pose in the world frame at the instant $t$ and $\mH_{o,t}^{w}$ the pose of the object in the world frame at the instant $t$. We define the \textbf{grasp-place pose similarity cost} as $c_{\textrm{grasp place similarity}}(\vtau) = d_{\textrm{SO(3)}  + \RR^3}(\mH_{ee,T/2}^{o}, \mH_{ee,T}^{o})$, that encourages the end effector pose w.r.t. the object frame to be the same in both the grasping moment and the placing moment.

\textbf{Exact weighting of cost terms}

In \Cref{tab:pick_reorient_weights} we present the weighting of the individual cost terms that we have used to obtain the trajectories for these scenarios of having to pickup a mug and reorienting it to fullfil a desired final pose.

\begin{table}[h]
\begin{center}
\scalebox{0.7}{
\begin{tabular}{c|c|c}
Description & Cost & Weight \\
\hline
\hline
Grasp pose evaluation & $E_{\vtheta}(\phi_{ee}(\vq_{t=T/2}), k)$ & $2.$\\
\hline
Trajectory smoothness & $c_{\textrm{smooth}}(\vtau)$ & $10.$\\
\hline
Table collision avoidance & $c_{\textrm{table coll.}}(\vtau)$ & $20.$\\
\hline
Initial configuration fixing cost &  $c_{fix}(\vtau)$ & $1.$\\
\hline
Pregrasping cost & $c_{\textrm{pregrasp}}(\vtau)$ & $1.$\\
\hline
Grasp-place pose similarity cost & $c_{\textrm{grasp place similarity}}(\vtau)$ & $10.$\\

\hline

\end{tabular}
}
\end{center}
\caption{This table summarizes the weighting of the individual cost terms that have been used for generating robot trajectories for the task of picking a mug in the first half of the trajectory and reorienting it to a desired final pose in the second half of the trajectory, as presented in \Cref{sec:exp_grasp_motion_opt} \& \Cref{app:pick_reorient}.}
\label{tab:pick_reorient_weights}
\end{table}

\subsubsection{Pick and place on shelves}
\label{app:pick_place_shelves}
In the following, we provide an extended presentation of the experiment of picking and placing on shelves. Similarly to the pick and reorient task, this experiment was chosen to evaluate the performance of \gls{se3dif} solving complex manipulation tasks jointly. This task is of high interest as both the set of affordable grasping poses and placing poses is very small due to the possible collisions with the shelves and therefore, jointly optimizing the trajectory and the grasp pose might be highly beneficial. The task is visualized in \Cref{fig:main_figure}. We evaluate the task similarly to the pick and reorient task. We generate 100 trajectories, execute them, and measure how close the grasped object is to the target pose in the last instant.

We consider the same objective function as for the pick and reorient task, but we also add a collision-avoidance cost to take the shelves collisions into account.

\textbf{Exact weighting of cost terms}

In \Cref{tab:pick_place_shelves_weights} we present the weighting of the individual cost terms that we have used to obtain the trajectories for these scenarios of picking a mug located inside a shelf and placing it in another desired final pose.

\begin{table}[h]
\begin{center}
\scalebox{0.7}{
\begin{tabular}{c|c|c}
Description & Cost & Weight \\
\hline
\hline
Grasp pose evaluation & $E_{\vtheta}(\phi_{ee}(\vq_{t=T/2}), k)$ & 1.\\
\hline
Trajectory smoothness & $c_{\textrm{smooth}}(\vtau)$ & 10.\\
\hline
Table collision avoidance & $c_{\textrm{table coll.}}(\vtau)$ & 10.\\
\hline
Initial configuration fixing cost &  $c_{fix}(\vtau)$ & 10.\\
\hline
Pregrasping cost & $c_{\textrm{pregrasp}}(\vtau)$ & 10.\\
\hline
Grasp-place pose similarity cost & $c_{\textrm{grasp place similarity}}(\vtau)$ & 1. \\
\hline
Box collision cost (shelf) & $c_{\textrm{box coll.}}(\vtau)$ & 10.\\

\hline

\end{tabular}
}
\end{center}
\caption{This table summarizes the weighting of the individual cost terms that have been used for generating robot trajectories for the task of picking and placing a mug inside a shelf as presented in \Cref{sec:exp_grasp_motion_opt} \& \Cref{app:pick_place_shelves}.}
\label{tab:pick_place_shelves_weights}
\end{table}

\subsection{Pointcloud based SE(3)-DiffusionFields}
\label{app:pointcloud}

\begin{figure*}[t]
	\centering
	\begin{minipage}{.90\textwidth}
		\includegraphics[width=.99\textwidth]{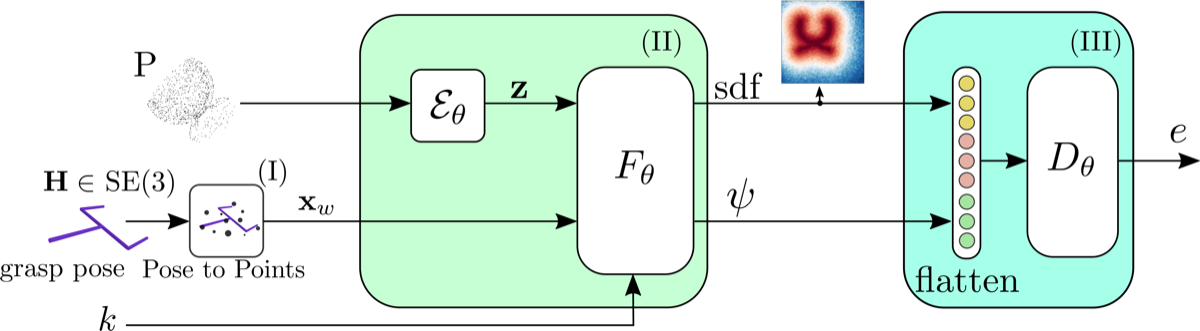}
	\end{minipage}
	\caption{An illustration of the PoiNt-\gls{se3dif} architecture for learning 6D grasp pose distributions. The architecture is similar to the autodecoder-based one (\cref{fig:architecture}) with a few modifications. We substitute the shape code embeddings and the object's pose transformation for a single pointcloud encoder $\gE_{\vtheta}$ that transforms an input pointcloud $P$ into a latent code $\vz$. We model $\gE_{\vtheta}$ with a \gls{vn}-PointNetI , that encodes SO(3) equivariant features.
	We train the model to jointly match the different object's SDF values (sdf) and to minimize the denoising loss. We jointly learn the parameters for the pointcloud encoder $\gE_{\vtheta}$, the features encoder $F_{\vtheta}$ and, the decoder $D_{\vtheta}$.}
	\label{fig:point_architecture}
\end{figure*}

The presented work is focused on evaluating the performance of diffusion models as both 6D grasp generative models and cost functions in trajectory optimization. Thus, to avoid perception related uncertainty, in this work, we assume the object shape and pose are known. We assume that we can rely on state-of-the-art object pose detection and
segmentation to estimate the object class and pose~\cite{xiang2017posecnn}. We apply an autodecoder approach~\cite{park2019deepsdf} and learn a set of latent codes $\vz$ that represent the different shapes. Then, in practice, given we know the exact object, we can retrieve the shape code $\vz$ given we know the index of the object.

For completeness, in this experimental section, we evaluate the performance of \gls{se3dif} with a Pointcloud encoder instead of an autodecoder. We modify the architecture in \Cref{fig:architecture} and add a pointcloud encoder $\gE_{\vtheta}$. We refer to this model as PoiNt-\gls{se3dif}. We present the modified architecture in \Cref{fig:point_architecture}.
We model the pointcloud encoder $\gE_{\vtheta}$ with a \gls{vn}-PointNet~\cite{deng2021vn}. The network outputs SO(3)-equivariant features that allow us to easily encode the orientation of the different objects. A similar network has been previously applied in \cite{xie2022neural} to learn the features of a graspable object.

We aim to evaluate the performance difference between the autodecoder based \gls{se3dif} and the pointcloud encoder based \gls{se3dif} models. We consider three scenarios for the autodecoder-based model: (i) Both object shape and pose are known, (ii) Only the shape is known and (iii) We don't know neither the shape nor the pose of the object. The case (i), where both pose and shape of the object are known, is presented in \cref{sec:exp_grasp_gen}. For the cases when either the object pose or both pose and shape code are unknown, we rely on pointclouds for inferring them. We follow a similar inference approach to the one proposed in \cite{park2019deepsdf} and extended it to infer also the pose of the object $\mH_o^w$. Given a pointcloud $P:\{\vx_n\}_{n=0}^N$ and the learned \gls{sdf} function $F_{\vtheta}^{sdf}$, we infer the $\mH_o^w$ by
\begin{align}
    \label{eq:rot_infer}
    \mH_{o}^{w*} = \argmin_{\mH_{o}^{w}} \frac{1}{N}\sum_{n=0}^N F_{\vtheta}^{sdf}(\mH_o^w \vx_n, \vz)
\end{align}
given $\vz$ the shape code of a known object. Intuitively, \eqref{eq:rot_infer} searches for the object pose $\mH_{o}^w$ that makes the pointcloud pose to match the one of the learned SDF function. We can think of this optimization problem as an \gls{icp} algorithm~\cite{chetverikov2002trimmed}, but instead of matching two sets of points, we match a set of points with the \gls{sdf} function. For the case when neither pose nor shape of the object is known, we infer both $\vz$ and $\mH_{o}^w$
\begin{align}
    \label{eq:latrot_infer}
    \mH_{o}^{w*}, \vz^* = \argmin_{\mH_{o}^{w}, \vz} \frac{1}{N}\sum_{n=0}^N F_{\vtheta}^{sdf}(\mH_o^w \vx_n, \vz) + \norm{\vz}^2,
\end{align}
and we extend the optimization for both the shape code $\vz$ and the object's pose $\mH_{o}^w$. We additionally add a L2 regularizer over $\vz$ as proposed by \cite{park2019deepsdf}. 

\begin{figure*}[t]
	\centering
	\begin{minipage}{.99\textwidth}
		\includegraphics[width=.99\textwidth]{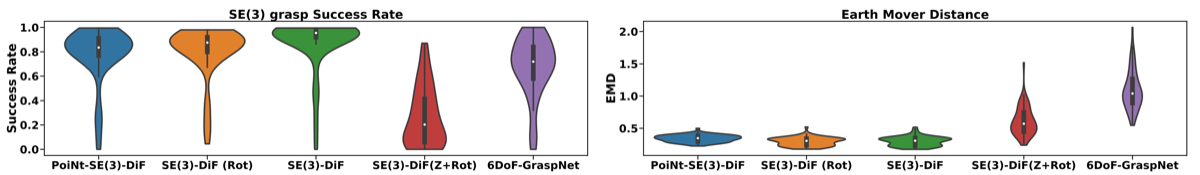}
	\end{minipage}
	\caption{Evaluation of the Success for picking with occlusions. PoiNt-\gls{se3dif} refers to the model with a pointcloud encoder, \gls{se3dif}~(Rot) to the model in which the pose is infer from the pointcloud, \gls{se3dif} the model in which both the pose and shape are known, \gls{se3dif}~(Z+Rot) the model in which both the object pose and shape codes are inferred from the pointcloud, and 6DoF-Graspnet~\cite{mousavian20196}.}
	\label{fig:results_point_grasps}
\end{figure*}

We evaluate the performance of the autodecoder-based approaches and the pointcloud encoder-based model w.r.t. their success rate in generating successful grasps and the \gls{emd}. We additionally add 6DoF-Graspnet~\cite{mousavian20196} as baseline to compare all the methods. We follow the same evaluation procedure from \Cref{sec:exp_grasp_gen}. We present the results of the evaluation in \Cref{fig:results_point_grasps}.
In \Cref{fig:results_point_grasps}, we name \gls{se3dif} the case where both object shape and pose are known, \gls{se3dif}~(Rot) the case where the object's shape is known, and the pose is inferred by pointclouds with \eqref{eq:rot_infer}, \gls{se3dif}~(Z+Rot) the case in which both object pose and shape are inferred with \eqref{eq:latrot_infer} and PoiNt-\gls{se3dif} the Pointcloud conditioned model. We observe a high performance in terms of both success rate and \gls{emd} for \gls{se3dif}, \gls{se3dif}~(Rot) and for the PoiNt-\gls{se3dif}. We also observe that the best success rate and \gls{emd} was achieved by \gls{se3dif}, followed by \gls{se3dif}~(Rot) and PoiNt-\gls{se3dif}. We hypothesize that this might be related to the unknown variables on each case. PoiNt-\gls{se3dif} needs to infer both the shape and the pose, while \gls{se3dif} assumes this to be known. We observe that \gls{se3dif}~(Z+Rot) was not able to achieve a high success rate. We were not able to properly infer both the shape code and the pose jointly by \eqref{eq:latrot_infer}. Therefore, if both the shape and the pose are unknown, we propose using PoiNt-\gls{se3dif}, while if the shape and the pose are known, we rather propose using the autodecoder approach. We observe, that all diffusion-based methods, except the \gls{se3dif}~(Z+Rot) outperformed 6DoF-GraspNet in terms of both success rate and Earth Mover Distance. While the success rate of 6DoF-GraspNet is is close to the one of the diffusion models, the \gls{emd} decays alot. This evaluation infers that the samples obtained by 6DoF-GraspNet are less diverse and the generation collapses to some modes in the dataset without covering it all.


\end{appendices}

\end{document}